\documentclass[journal]{IEEEtran}
\usepackage{makeidx}  
\usepackage{setspace}
\usepackage{graphicx}

\usepackage{amsmath,amsfonts,amssymb}
\usepackage{mathabx}
\usepackage{bbm}
\usepackage{color}
\usepackage{wrapfig}
\usepackage{todonotes}
\usepackage[font=footnotesize,labelfont=bf]{caption}
\usepackage{hyperref}
\usepackage{colortbl}

\definecolor{blue}{rgb}{0,0,0}

\definecolor{anti-flashwhite}{rgb}{0.95, 0.95, 0.96}

\renewcommand{\boldsymbol}{}

\newcommand{\bC}{\boldsymbol{C}}
\newcommand{\bS}{\boldsymbol{S}}

\newcommand{\bU}{\boldsymbol{U}}

\newcommand{\bI}{\boldsymbol{I}}

\newcommand{\bb}{\boldsymbol{b}}
\newcommand{\bpi}{\boldsymbol{\pi}}
\newcommand{\bmu}{\boldsymbol{\mu}}
\newcommand{\bsigma}{\boldsymbol{\sigma}}
\newcommand{\bSigma}{\boldsymbol{\Sigma}}

\newcommand{\bLambda}{\boldsymbol{\Lambda}}
\newcommand{\btheta}{\boldsymbol{\theta}}

\newcommand{\bepsilon}{\boldsymbol{\epsilon}}

\newcommand{\indep}{\rotatebox[origin=c]{90}{$\models$}}

\newcommand{\by}{\boldsymbol{y}}

\newcommand{\bx}{\boldsymbol{x}}

\newcommand{\bY}{\boldsymbol{Y}}
\newcommand{\bZ}{\boldsymbol{Z}}

\newcommand{\bV}{\boldsymbol{V}}

\newcommand{\bW}{\boldsymbol{W}}

\newcommand{\delequal}{\overset{\Delta}{=}}

\newcommand{\obs}{\mbox{\tiny$\mathcal{O}$}}
\newcommand{\misi}{\mbox{\tiny$\mathcal{M}_i$}}
\newcommand{\obsi}{\mbox{\tiny$\mathcal{O}_i$}}

\newcommand{\obsj}{\mbox{\tiny$\mathcal{P}_j$}}

\usepackage{soul}

\newcommand{\Expect}{{\rm I\kern-.3em E}}

\newcommand{\figdir}{figures}

\definecolor{very-light-gray}{gray}{0.90}

\ifCLASSINFOpdf
\else
\fi
\ifCLASSOPTIONcompsoc
  \usepackage[caption=false,font=normalsize,labelfont=sf,textfont=sf]{subfig}
\else
  \usepackage[caption=false,font=footnotesize]{subfig}
\fi
\begin{document}
%
\title{Medical Image Imputation from Image Collections}
%
%
%
%

\author{Adrian V. Dalca,
	Katherine L. Bouman,
	William T. Freeman, 
	Natalia S. Rost, 
	Mert R. Sabuncu, 
	Polina Golland \\
	for the Alzheimer’s Disease Neuroimaging Initiative*
\thanks{Adrian V. Dalca is with the Computer Science and Artificial Intelligence Lab, MIT (main contact: adalca@csail.mit.edu) and also Martinos Center for Biomedical Imaging, Massachusetts General Hospital, HMS.}
\thanks{Katherine L. Bouman and Polina Golland are with the Computer Science and Artificial Intelligence Lab, MIT.}
\thanks{William T. Freeman is with the Computer Science and Artificial Intelligence Lab, MIT and Google.}
\thanks{Mert R. Sabuncu is with the the School of Electrical and Computer Engineering, and Meinig School of Biomedical Engineering, Cornell University.}
\thanks{Natalia S. Rost is with the Department of Neurology, Massachusetts General Hospital, HMS.}
\thanks{*Data used in preparation of this article were obtained from the Alzheimer’s Disease
	Neuroimaging Initiative (ADNI) database (adni.loni.usc.edu). As such, the investigators
	within the ADNI contributed to the design and implementation of ADNI and/or provided data
	but did not participate in analysis or writing of this report. A complete listing of ADNI
	investigators can be found at:
	\url{http://adni.loni.usc.edu/wp-content/uploads/how_to_apply/ADNI_Acknowledgement_List.pdf}}
}

\maketitle

\begin{abstract}
We  present an algorithm for creating high resolution anatomically plausible images consistent with acquired 
clinical brain MRI scans with large inter-slice spacing. Although large data sets of clinical images contain a wealth of information, time constraints during acquisition result in sparse scans that fail to capture much of the anatomy. These characteristics often render computational analysis impractical as many image analysis algorithms tend to fail when applied to such images. Highly specialized algorithms that explicitly handle sparse slice spacing do not generalize well across problem domains. In contrast, we aim to enable application of existing algorithms that were originally developed for high resolution research scans to significantly undersampled scans.  We introduce a generative model that captures fine-scale anatomical structure across subjects in clinical image collections and derive an algorithm for filling in the missing data in scans with large inter-slice spacing. {\color{blue}Our experimental results demonstrate that the resulting method outperforms state-of-the-art upsampling super-resolution techniques, and promises to facilitate subsequent analysis not previously possible with scans of this quality.} Our implementation is freely available at \url{https://github.com/adalca/papago}. 
\end{abstract}


%
\IEEEpeerreviewmaketitle

\section{Introduction}
\label{sec:local}

Increasingly open image acquisition efforts in clinical practice are driving dramatic increases in the number and size of patient cohorts in clinical archives. Unfortunately, clinical scans are typically of dramatically lower resolution than the research scans that motivate most methodological development. Specifically, while slice \textit{thickness} can vary depending on the clinical study or scan, inter-slice \textit{spacing} is often significantly larger than the in-plane resolution of individual slices. This results in missing voxels that are typically filled via interpolation. 

 %


\begin{figure}[t]
	\centering
	\includegraphics[width=1\linewidth]{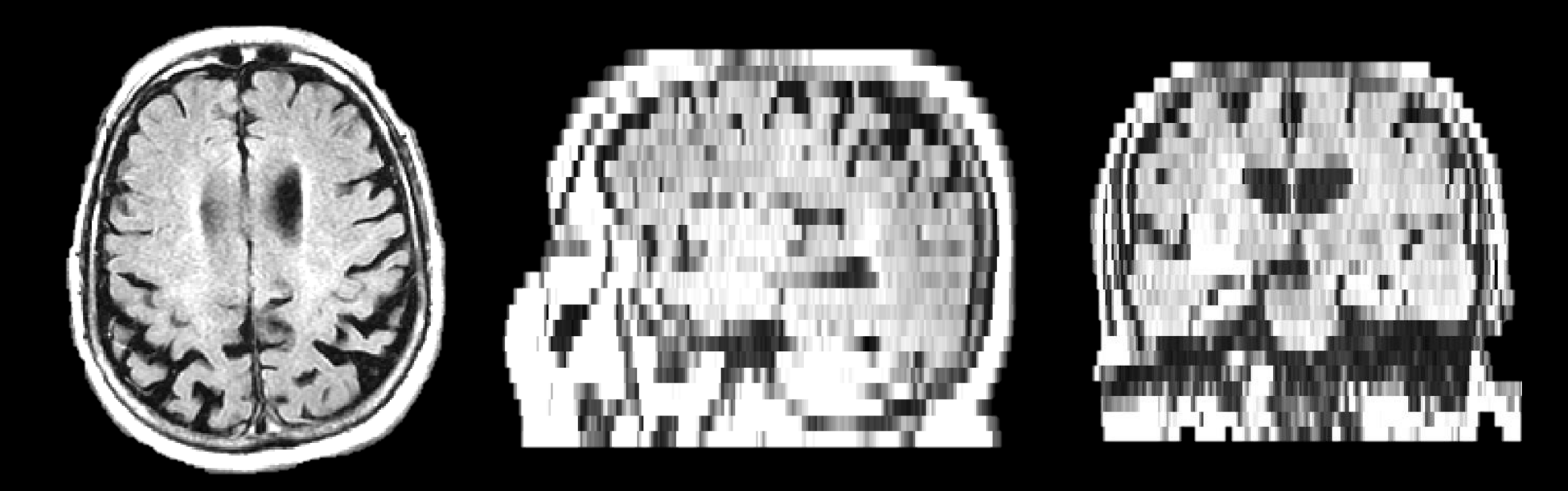}
	\caption{
	\textbf{An example scan from our clinical dataset.} The three panels display axial, sagittal and coronal slices, respectively. 
		While axial in-plane resolution can be similar to that of a research scan, slice spacing is significantly larger. We visualize the saggital and coronal views using nearest neighbor interpolation.
	}
	\label{fig:strokeexample}
\end{figure}

Our work is motivated by a study that includes 
brain MRI scans of thousands of stroke patients acquired within 48 hours of stroke onset. The study aims to quantify white matter disease burden~\cite{rost2010}, 
necessitating skull stripping and deformable registration into a common coordinate frame~\cite{schmidt2012,sridharan2013,vanleemput2001}. 
The volumes are severely under-sampled ($0.85$mm~$\times$~$0.85$mm~$\times$~$6$mm) due to constraints of acute stroke care (Fig.~\ref{fig:strokeexample}). Such undersampling is typical of modalities, such as T2-FLAIR, that aim to characterize tissue properties, even in research studies like ADNI~\cite{jack2008}.

In undersampled scans, the image is no longer smooth, and the anatomical structure may change substantially between consecutive slices (Fig.~\ref{fig:strokeexample}). Since such clinically acquired scans violate underlying assumptions of many algorithms, even basic tasks such as skull stripping and deformable registration present significant challenges, yet are often necessary for downstream analysis~\cite{dalca2016patch,dimartino2014,hill2001,jack2008,rost2010,shi2013,sridharan2013}. 

We present a novel method for constructing high resolution anatomically plausible volumetric images consistent with the available slices in sparsely sampled clinical scans.  Importantly, our method does not require any high resolution scans or expert annotations for training. It instead imputes the missing structure by learning solely from the available collection of sparsely sampled clinical scans. 
The restored images represent plausible anatomy. They promise to act as a medium for enabling computational analysis of clinical scans with existing techniques originally developed for high resolution, isotropic research scans.  For example, although imputed data should not be used in clinical evaluation, the brain mask  obtained through skull stripping of the restored scan can be applied to the original clinical scan to improve subsequent analyses.



\subsection{Prior Work} 


Many image restoration techniques depend on having enough information in a single image to synthesize data. Traditional interpolation methods, such as linear, cubic or spline~\cite{schoenberg1973}, assume a functional representation of the image. They treat the low resolution voxels as samples, or observations, and estimate function parameters to infer missing voxel values. 
Patch-based superresolution algorithms use fine-scale redundancy within a single scan~\cite{freeman2002,glasner2009,manjon2010,manjon2012,plenge2013}. The key idea is to fill in the missing details by identifying similar image patches in the same image that might contain relevant detail~\cite{manjon2010,plenge2013}. This approach depends on having enough repetitive detail in a scan to capture and re-synthesize high frequency information. Unfortunately, clinical images are often characterized by sampling that is too sparse to adequately fit functional representations or provide enough fine-scale information to recover the lost detail. For example,~$6$mm slice spacing, typical of many clinical scans including our motivating example, is far too high to accurately estimate approximating functions without prior knowledge. In such cases, a single image is unlikely to contain enough fine-scale information to provide anatomically plausible reconstructions in the direction of slice acquisition, as we demonstrate later in the paper.

Alternatively, one can use additional data to synthesize better images. Many superresolution  algorithms use multiple scans of the same subject, such as multiple low resolution acquisitions with small shift differences to synthesize a single volume~\cite{carmi2006,jog2014,plenge2013}. However, such acquisitions are not commonly available in the clinical setting.

Nonparametric and convolutional neural-network (CNN) based upsampling methods that tackle the problem of superresolution often rely on an external dataset of high resolution data or cannot handle extreme undersampling present in clinical scans.  
For example, some methods fill in missing data by matching a low resolution image patch from the input scan with a high resolution image patch from the training dataset~\cite{coupe2011,iglesias2013,jog2014,konukoglu2013,rousseau2010non,rousseau2011}. 
Similarly, CNN-based upsampling methods approximate completion functions, but require high resolution scans for training~\cite{dong2014,oktay2016}. 
%
%
A recent approach to improve resolution from a collection of scans with sparse slices jointly upsamples all images using non-local means~\cite{rousseau2010}. However this method has only been demonstrated on slice spacing of roughly three times the in-plane resolution, and in our experience similar non-parametric methods fail to upsample clinical scans with more significant undersampling. 

\begin{figure*}[t]
	\centering
	\includegraphics[width=0.9\linewidth]{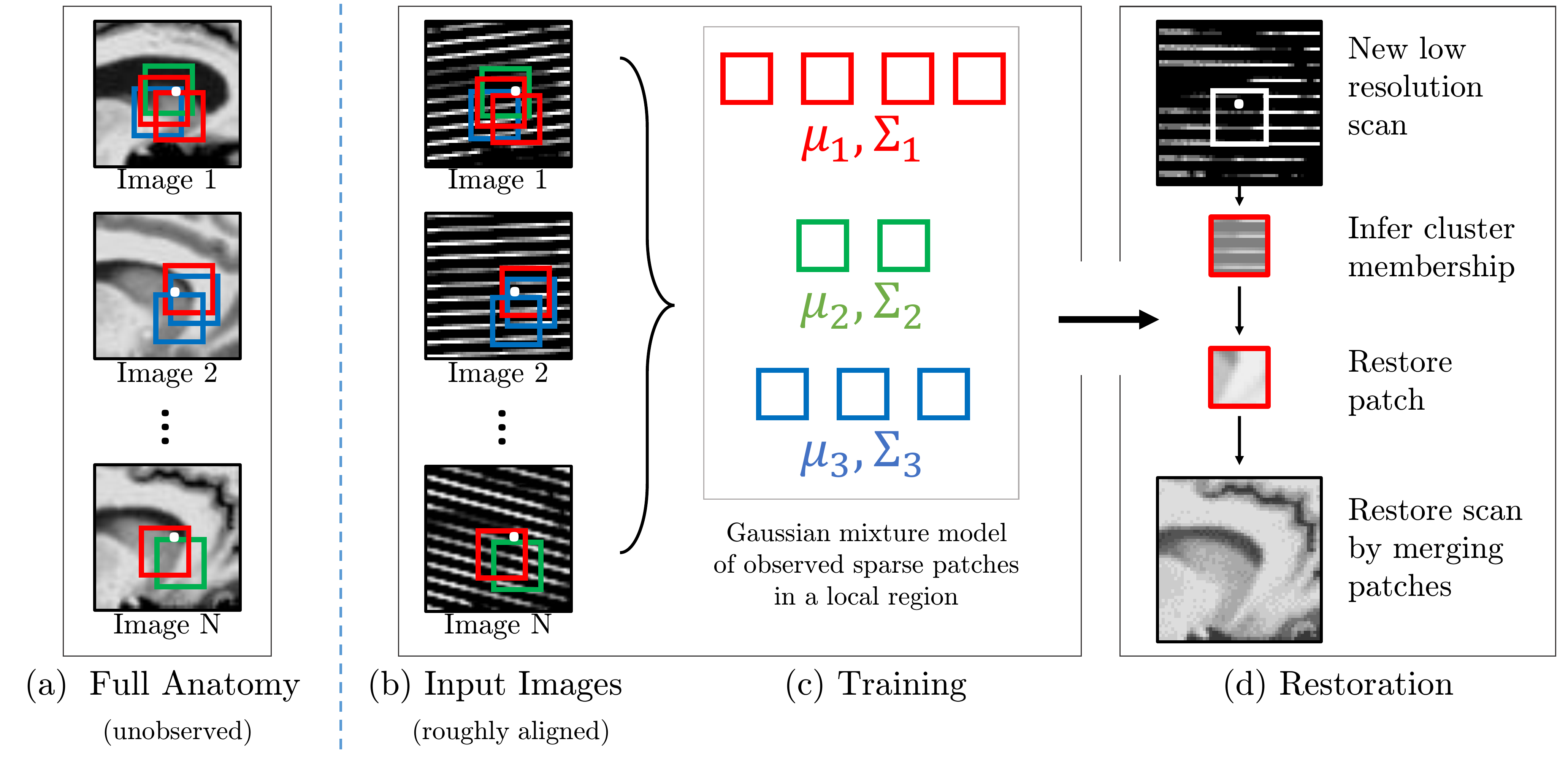}
	\vspace{-0.05cm}
	\caption{\textbf{Image imputation for a subvolume.} (a) Full resolution images, shown for illustration only. These are unobserved by the algorithm. (b) Sparse planes acquired in clinical scans. (c)~During learning, we train a GMM that captures the low dimensional nature of patch variability in a region around a particular location (white dot). (d) Given a sparsely sampled scan, we infer the most likely cluster for each 3D patch, and restore the missing data using the learned model and the observed voxels. We form the final volume from overlapping restored patches. 2D images are shown for illustration only, the algorithms operate fully in 3D.}
	\label{fig:localmodels-method}
\end{figure*}

Our work relies on a low dimensional embedding of image patches with missing voxels.
Parametric patch methods and low dimensional embeddings have been used to model the common structure of image patches from full resolution images, but are typically not designed to handle missing data. Specifically, priors~\cite{yang2010image} and Gaussian Mixture Models~\cite{zoran2011learning,zoran2012} have been used in both medical and natural images for classification~\cite{bhatia2014} and denoising~\cite{elad2006image,zoran2012}.
The procedures used for training of these models rely on having full resolution patches with no missing data in the training phase. 

Unfortunately, high (full) resolution training datasets are not readily available for many image contrasts and scanners, and may not adequately represent pathology or other properties of clinical populations.
Acquiring the appropriate high resolution training image data is often infeasible, and here we explicitly focus on the realistic clinical scenario where only sparsely sampled images are available.




\subsection{Method Overview} 
We take advantage of the fact that local fine scale structure is shared in a population of medical images, and each scan with sparse slices captures some partial aspect of this structure. We borrow ideas from Gaussian Mixture Model (GMM) for image patch priors~\cite{zoran2012}, low dimensional Gaussian embeddings~\cite{ilin2010,zhang2013gaussian}, and missing data models~\cite{ilin2010,little2014} to develop a probabilistic generative model for sparse 3D image patches around a particular location
using a low-dimensional GMM with partial observations. We derive the EM algorithm for maximum likelihood estimation of the model parameters and discuss related modeling choices. Given a new sparsely sampled scan, the maximum \textit{a posteriori} estimate of the latent structure yields the imputed high resolution image. We evaluate our algorithm using scans from the ADNI cohort, and demonstrate its utility in the context of the motivating stroke study. We investigate the behaviour of our model under different parameter settings, and illustrate an example of potential improvements in the downstream analysis using an example task of skull stripping.

%



{\color{blue} This paper extends the preliminary version of the method	presented at the 2017 Conference on Information Processing in Medical Imaging~\cite{dalca2017}. Here, we improve model inference by removing parameter co-dependency between iterations and providing new parameter initialization. We provide detailed derivations and discuss an alternative related model. Finally, we provide an analysis of important model parameters, present results for more subjects, and illustrate more example reconstructions.}
The paper is organized as follows. Section~\ref{sec:mixture-full} introduces the model and learning algorithm. Section~\ref{sec:implementation} discusses implementation details. We present experiments and analysis of the algorithm's behavior in Section~\ref{sec:experiments}. We discuss important modeling aspects and related models in Section~\ref{sec:discussion}. We include an Appendix and Supplementary Material with detailed derivations of the EM algorithm for the proposed models.



\section{Method}
\label{sec:mixture-full}
\label{sec:loc-model}


In this section, we construct a generative model for sparse image patches, present the resulting learning algorithm, and describe our image restoration procedure. 

{\color{blue}Let~$\{\bY_1,...,\bY_N\}$ be a collection of scans with large inter-slice spaces, \textit{roughly} aligned into a common atlas space (we use affine transformations in our experiments).} For each image~$\bY_i$ in the collection, only a few slices are observed. We seek to restore an anatomically plausible high resolution volume by imputing the missing voxel values.

{\color{blue}We capture local structure using image patches. We assume a constant patch shape, and in our experiments use a 3D~$11$x$11$x$11$ shape. We use~$\by_i$  
to denote a~$D$-length vector that contains voxels of the image patch centered at a certain location in image~$\bY_i$.}
 We perform inference at each location independently and stitch the results into the final image as described later in this section.
Fig.~\ref{fig:localmodels-method} provides an overview of the method.

\subsection{Generative Model}
\label{sec:full-res-model}

%

\begin{figure}[t]
	\centering
	\includegraphics[width=0.7\linewidth]{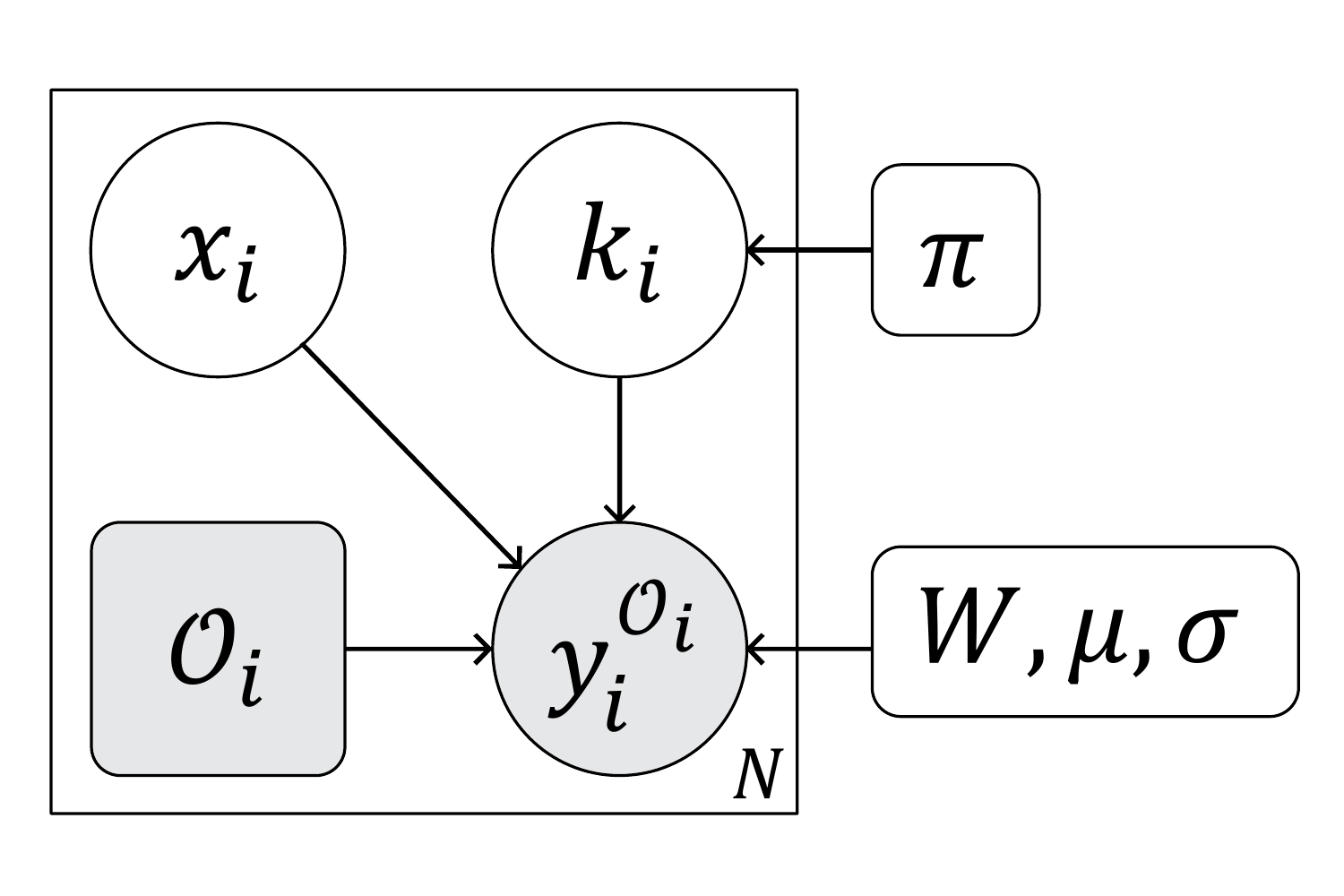}
	\vspace{-0.25cm}
	\caption{\textbf{Graphical representation of our model.} Circles indicate random variables and rounded squares represent parameters. Shading represents observed quantities and the plate indicates replication. 
		The observed patch voxels~$\by_i^{\obsi}$ form a subset of patch~$\by_i$ extracted by the mask~$\mathcal{O}_i$ and are generated from a multivariate Gaussian distribution conditioned on the latent cluster~$k_i$ and the  latent patch representation~$\bx_i$. Parameters~$\bmu$ and~$\bW$ define the mean and the variance of the Gaussian components of the mixture, and~$\bsigma^2$ is the image noise variance. \vspace{-0.4cm}}
	\label{fig:localmodels-graphical-model}
\end{figure}

We treat an image patch as a high dimensional manifestation of a low dimensional representation, with the intuition that the covariation within image patches has small intrinsic dimensionality relative to the number of voxels in the patch. To capture the anatomical variability across subjects, we employ a Gaussian Mixture Model (GMM) to represent local structure of 3D patches in the vicinity of a particular location across the entire collection. We then explicitly model the observed and missing information. Fig~\ref{fig:localmodels-graphical-model} presents the corresponding graphical model. 


{\color{blue}
	We model the latent low dimensional patch representation~$\bx_i$ of length~$d < D$ as a normal random variable
	\begin{align}
	\bx_i &\sim \mathcal{N}(0, \bI_{d \times d}),
	\label{eq:latent_x}
	\end{align}
	where~$\mathcal{N}(\mu, \Sigma)$ denotes the multivariate Gaussian distribution with mean~$\mu$ and covariance~$\Sigma$. 
	We draw latent cluster assignment~$k$ from a categorical distribution defined by a length-$K$ vector~$\pi$ of cluster probabilities, and treat image patch~$\by_i$ as a high dimensional observation of~$\bx_i$ drawn from a~$K$-component multivariate GMM. Specifically, conditioned on the drawn cluster~$k$,
	\begin{align}
	\by_i &= \bmu_k + \bW_k \bx_i + \bepsilon_{i}, \quad \mbox{where}   \\
	\bepsilon_i &\sim \mathcal{N}(0, \sigma^2_k \bI_{D \times D}), \quad \mbox{and} \quad \bepsilon_{i} \hspace{0.01cm}  \hspace{0.04cm} \indep \hspace{0.05cm} \bx_i. \nonumber
	\label{eq:gmm-pca}
	\end{align}
	Vector~$\mu_k$ is the patch mean of cluster~$k$, matrix~$\bW_k$ shapes the covariance structure of~$\by_i$, and~$\sigma^2_k$ is the variance of image noise. This model implies $\Expect[y_i|k] = \mu_k$ and \mbox{$\bC_k \delequal \Expect[(y_i - \mu_k)(y_j-\mu_k)^T | k] = \bW_k \bW_k^T + \sigma_k^2\bI_{D \times D}$}. 
	
	Defining~$\theta = \{\mu_k, W_k, \sigma^2_k, \pi_k\}_{k=1}^K$, the likelihood of all patches~$\mathcal{Y} = \{\by_i\}$ at this location under the mixture model is
	\begin{align}
	p(\mathcal{Y}; \theta)
	= \prod_i \sum_k \pi_k \mathcal{N}(\by_i; \bmu_k, \bC_k).
	\end{align}
}

In our clinical images, only a few slices are known.  To model sparse observations, we let~$\mathcal{O}_i$ be the set of observed voxels in patch~$\by_i$, and~$\by_i^{\obsi}$ be the corresponding vector of their intensity values:
%
%
\begin{align}
\by_i^{\obsi} &= \bmu_{k}^{\obsi} + \bW_{k}^{\obsi} \bx_i + \bepsilon_{i}^{\obsi} ,
\end{align}
where~$\bW_{k}^{\obsi}$ comprises rows of~$\bW_k$ that correspond to the observed voxel set~$\mathcal{O}_i$. 
The likelihood of the \textit{observed} data~\mbox{$\mathcal{Y^{\obs}}=\{\by^{\obsi}_i\}$} is therefore
\begin{align}
p(\mathcal{Y}^{\obs}; \theta) &= \prod_i \sum_k \pi_k \mathcal{N}(\by_i^{\obsi}; \bmu_k^{\obsi}, \bC_k^{\obsi\obsi}),
\label{eq:local-gmmppca-likelihood}
\end{align}
%
where matrix~$\bC_{k}^{\obsi\obsi}$ extracts the rows and columns of~$\bC_k$ that correspond to the observed voxel subset~$\mathcal{O}_i$. 

We do not explicitly model slice thickness, as in many clinical datasets this thickness is unknown or varies by site, scanner or acquisition. Instead, we simply treat the original data as high resolution thin planes and analyze the effects of varying slice thickness on the results in the experimental evaluation of the method.

We also investigated an alternative modeling choice where each missing voxel of patch~$\by_i$ is modelled as a latent variable. This assumption can optionally be combined with the latent low-dimensional patch representation. We discuss this alternative choice in Section~\ref{sec:discussion}, and provide parameter updates in the Supplementary Material. Unfortunately, the resulting algorithm is prohibitively slow.

\subsection{Learning}

Given a collection of observed patches~$\mathcal{Y}^{\obs}$, we seek the maximum likelihood estimates of the model parameters~$\{\bmu_k, \bW_k, \sigma^2_k\}$ and~$\bpi$ under the likelihood~\eqref{eq:local-gmmppca-likelihood}. 
%
%
%
We derive the Expectation Maximization algorithm~\cite{dempster1977} in Appendix~\ref{app:em}, and present the update equations and their interpretations below. 

The \textbf{expectation step} updates the class memberships:
\begin{align}
\gamma_{ik} &\delequal p(k | \by_i^{\obsi};  \theta) \nonumber \\
&= \frac{\pi_k \mathcal{N}(\by_i^{\obsi};\bmu_{k}^{\obsi},\bC_{k}^{\obsi\obsi})}{\sum_{k'} \pi_{k'} \mathcal{N}(\by_i^{\obsi};\bmu_{k'}^{\obsi},\bC_{k'}^{\obsi\obsi})},
\label{eq:local-gmmppca-estep-mem}
\end{align}
and the statistics of the low dimensional representation~$\bx_i$ for each image patch~$\by^{\obsi}_i$ as "explained" by cluster~$k$: 
\begin{align}
\widehat{\bx}_{ik} &\delequal \Expect[\bx_i|k]  \label{eq:local-gmmppca-estep-x}  \\
&= \left( (\bW_{k}^{\obsi})^T  (\bW_{k}^{\obsi}) + \sigma_k^{2} \bI_{d \times d} \right)^{-1}  (\bW_k^{\obsi})^{T} (\by_i^{\obsi} - \bmu_k^{\obsi}),
\nonumber \\
\bS_{ik} &\delequal \Expect[\bx_i\bx^T_i|k] - \widehat{\bx}_{ik} \widehat{\bx}_{ik}^T \nonumber \\
&= \sigma_k^2 \left( (\bW_{k}^{\obsi})^T (\bW_{k}^{\obsi}) + \sigma_k^{2} \bI_{d \times d} \right)^{-1}.
\label{eq:local-gmmppca-estep-S} 
\end{align}
%
%
%
%
%
We let~$\mathcal{P}_j$ be the set of patches in which voxel~$j$ is observed, and form the following normalized mean statistics:
\begin{align}
\delta_{ik} &= \frac{\gamma_{ik}}{\sum_{i'\in\obsj} \gamma_{i'k}}  \label{eq:mean-exp-delta} \\
b_j &= \sum_{i\in\obsj} \delta_{ik} \Expect[x_i|k] 		= \sum_{i\in\obsj} \delta_{ik} \widehat{x}_{ik} \label{eq:mean-exp-x}  \\
A_j &= \sum_{i\in\obsj} \delta_{ik} \Expect[x_ix_i^T|k]	= \sum_{i\in\obsj} \delta_{ik} (\widehat{\bx}_{ik} \widehat{\bx}_{ik}^{T} + \bS_{ik}).  \label{eq:mean-exp-cov}
\end{align}

The \textbf{maximization step} uses the observed voxels to update the model parameters. We let~$\by_i^{j}$ be the~$j\textsuperscript{th}$ element of vector~$\by_i$, and update the cluster mean as a convex combination of observed voxels:
\begin{align}
\bmu_k^{j} &\leftarrow \frac{\sum_{i\in\obsj} \gamma_{ik} (1 - \widehat{x}_i^T A_j^{-1} b_j ) y_{i}^{j}}{\sum_{i'\in\obsj} \gamma_{i'k} (1 - \widehat{x}_{i'}^T A_j^{-1} b_j )  } .
\end{align}
%
The covariance factors and image noise variance are updated based on the statistics of the low dimensional representation from~\eqref{eq:mean-exp-x} and~\eqref{eq:mean-exp-cov}:
\begin{align}
\bW_{k}^j &\leftarrow \sum_{i\in\obsj} \delta_{ik} (y_{i}^{j} - \mu_{k}^{j}) \widehat{\bx}_{ik}^{T} A_j^{-1}, \\
\sigma_k^{2} &\leftarrow \frac{\sum_j\sum_{i\in\obsj} \gamma_{ik} \left[ (y_{i}^{j} - \mu_{k}^{j} - \bW_{k}^j\widehat{\bx}_{ik})^2 + \bW_{k}^j \bS_{ik} (\bW_{k}^j)^{T} \right]}{\sum_j\sum_{i'\in\obsj} \gamma_{i'k}}.
\label{eq:local-gmmppca-mstep}
\end{align}
where~$W_k^j$ is the~$j\textsuperscript{th}$ row of matrix~$W_k$. Finally, we update the cluster proportions:
\begin{align}
\pi_k &= \frac{1}{N} \sum_i \gamma_{ik}.
\end{align}
%

Intuitively, learning our model with sparse data is possible because each image patch provides a slightly different subset of voxel observations that contribute to the parameter estimation (Fig.~\ref{fig:localmodels-method}). In our experiments, all subject scans have the same acquisition direction. Despite different affine transformations to the atlas space for each subject, some voxel \textit{pairs} are still never observed in the same patch, 
resulting in missing entries of the covariance matrix. Using a low-rank approximation for the covariance matrix regularized the estimates.

{\color{blue}
	Upon convergence of the EM updates, we compute the cluster covariance~\mbox{$\bC_k = \bW_k \bW_k^T + \sigma_k^2\bI_{D \times D}$} for each~$k$.
}

\subsection{Imputation}

{\color{blue}
To restore an individual patch~$\by_i$, we compute the \textit{maximum-a-posteriori} (MAP) estimate of the image patch:
\begin{align}
\hat{\by}_i &= \arg\max_{\by_i} p({\by_i} | \by_i^{\obsi} ; \btheta) \nonumber \\  
&= \arg\max_{\by_i} \sum_k p(k | \by_i^{\obsi})  \int_{\bx_i}  p(\by_i|\bx_i) p(\bx_i|k,\by_i^{\obsi}) dx_i \nonumber \\
&= \arg\max_{\by_i} \sum_k \gamma_{ik} \int_{\bx_i}  p(\by_i|\bx_i) p(\bx_i|{k},\by_i^{\obsi}) dx_i \nonumber \\
&= \arg\max_{\by_i} \sum_k \gamma_{ik} \mathcal{N}(y_i ; \mu_{{k}} + W_{{k}} \widehat{x}_{i{k}} ,  \bSigma_{i{k}}), \nonumber 
\end{align}
where~$\bSigma_{i{k}} = \sigma_{{k}}^2 I_{D \times D} + \bW_{{k}} (S_{i{k}} + \widehat{\bx}_{ik} \widehat{\bx}_{ik}^T) \bW_{{k}}^T$. Due to the high-dimensional nature of the data, most cluster membership estimates are very close to~$0$ or~$1$. We therefore first estimate the most likely cluster~$\widehat{k}$ for patch~$\by_i$ by selecting the cluster with the highest membership~$\gamma_{ik}$. We estimate the low dimensional representation~$\widehat{\bx}_{i\widehat{k}}$ given the observed voxels~$\by_i^{\obsi}$ using (7), which yields the high resolution imputed patch:
\begin{align}
\hat{\by}_i &= \bmu_{\widehat{k}} + \bW_{\widehat{k}} \widehat{\bx}_{i\widehat{k}}.
\label{eq:restoration-LSR}
\end{align}

By restoring the scans using this MAP solution, we perform \textit{conditional mean imputation}~(c.f.~17, Sec.4.2.2), and demonstrate the reconstructions in our experiments. In addition, our model enables imputation of each patch by sampling the posterior~$p( \cdot | \by_i^{\obsi} ; \btheta) \approx \mathcal{N}(\cdot ; \mu_{\widehat{k}} + W_{\widehat{k}} \widehat{x}_{i\widehat{k}} ,  \bSigma_{i\widehat{k}})$, providing a better estimation of the residual noise.  Depending on the desired downstream application, sampling-based imputation may be desired.

We average overlapping restored patches using standard techniques~[18] to form the restored volume. 
}

\begin{figure*}[!t]
	\centering
	\includegraphics[width=1\linewidth]{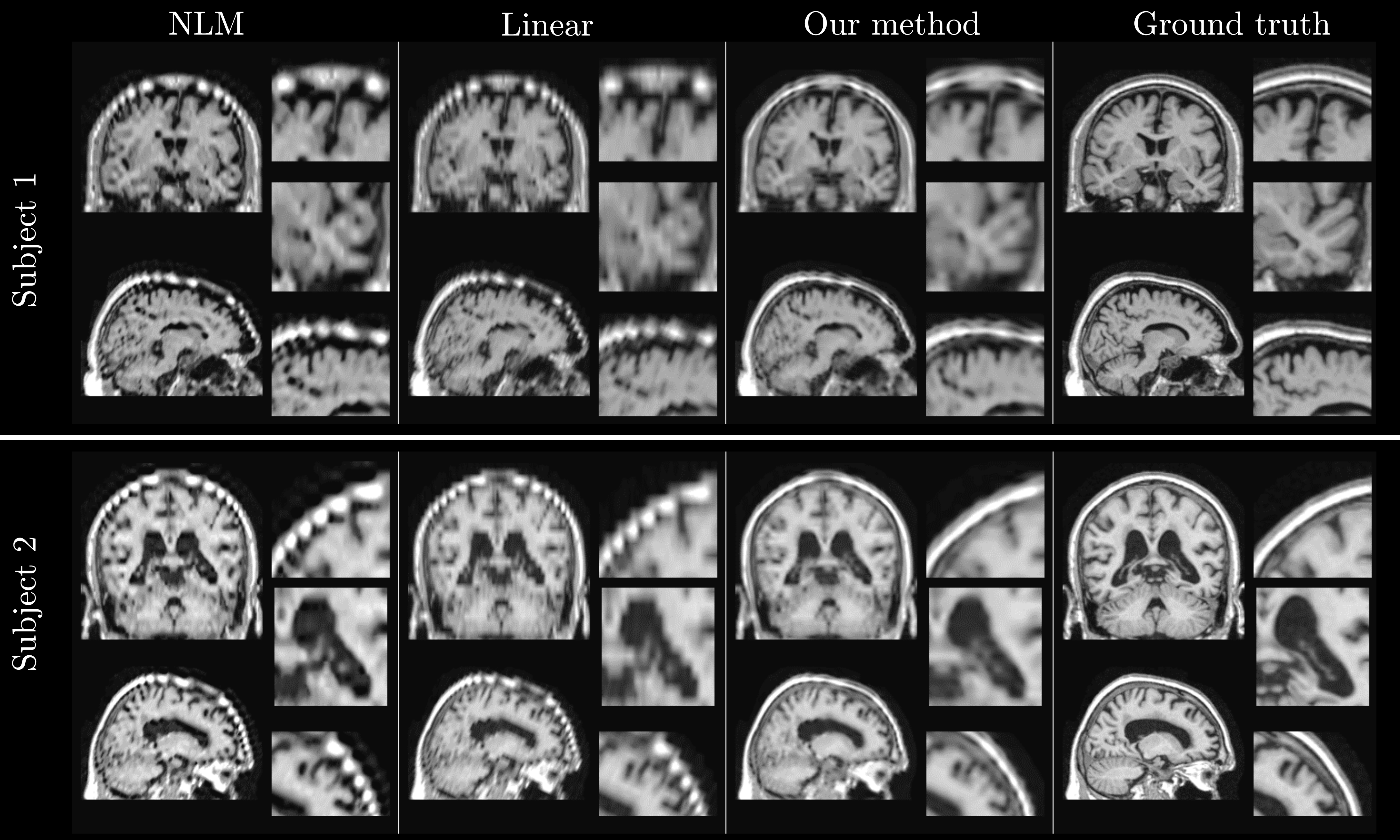}
	\caption{\textbf{Representative restorations in the ADNI dataset}. Reconstruction by NLM, linear interpolation, and our method, and the original high resolution images for two representative subjects in the study. Our method reconstructs more anatomically plausible substructures as can be especially seen in the close-up panels of the skull, ventricles, and temporal lobe. Additional examples are available in the Supplementary Materials. \vspace{-0.5cm}}
	\label{fig:buckner-images}
\end{figure*}

\section{Implementation}
\label{sec:implementation}

We work in the atlas space, and approximate voxels as either observed or missing in this space by thresholding interpolation weights. To limit interpolation effects due to affine alignment on the results, we set a higher threshold for regions with high image gradients than in regions with low gradients. Parameter estimation could be implemented to include transformation of the model parameters into the subject-specific space in order to optimally use the observed voxels, but this leads to computationally prohibitive updates. 

{\color{blue} We stack together the affinely registered sparse images from the entire collection. We learn a single set of mixture model parameters within overlapping subvolumes of~$21\times21\times21$ voxels in the isotropically sampled common atlas space. Subvolumes are centered 11 voxels apart in each direction.  We use a cubic patch of size~$11 \times11  \times11$ voxels, and instead of selecting just one patch from each volume at a given location, we collect all overlapping patches within the subvolume centered at that location. This aggregation provides more data for each model, which is crucial when working with severely undersampled volumes. Moreover, including nearby voxels offers robustness in the face of image misalignment. Given the learned parameters at each location, we restore all overlapping patches within a subvolume. }


While learning is performed in the common atlas space, we restore each volume in its original image space to limit the effects of interpolation. Specifically, we apply the inverse of the estimated subject-specific affine transformation to the cluster statistics prior to performing subject-specific inference. 


Our implementation is freely available at \url{https://github.com/adalca/papago}.

\section{Experiments}
\label{sec:experiments}

We demonstrate the proposed imputation algorithm on two datasets and evaluate the results both visually and quantitatively. We also include an example of how imputation can aid in a skull stripping task. 

\subsection{Data: ADNI dataset}

\enskip We evaluate our algorithm using 826 T1-weighted brain MR images from ADNI~\cite{jack2008}~\footnote{Data used in the preparation of this article were obtained from the Alzheimer’s Disease Neuroimaging Initiative (ADNI) database (adni.loni.usc.edu). The primary goal of ADNI has been to test whether serial magnetic resonance imaging (MRI), positron emission tomography (PET), other biological markers, and clinical and neuropsychological assessment can be combined to measure the progression of mild cognitive impairment (MCI) and early Alzheimer’s disease (AD).}. 
We downsample the isotropic $1$mm$^3$ images to slice separation of $6$mm ($1$mm~$\times$~$1$mm~in-plane) in the axial direction to be of comparable quality with the clinical dataset. We use these low resolution images as input. All downsampled scans are affinely registered to a T1 atlas. The original images serve as the ground truth for quantitative comparisons. After learning model parameters using the data set, we evaluate the quality of the resulting imputations.

\subsection{Evaluation}
%
%
%
We compare our algorithm to three upsampling methods: nearest neighbour~(NN) interpolation, non-local means (NLM) upsampling, and linear interpolation~\cite{manjon2010}.
We compare the reconstructed images to the original isotropic volumes both visually and quantitatively. We use the mean squared error, 
\begin{align}
\mbox{MSE}~(Z,Z_o)=\frac{1}{N}\sum || Z-Z_o ||^2,
\end{align} 
of the reconstructed image~$\bZ$ relative to the original high resolution scan~$\bZ_o$. We also compute the related peak signal to noise ratio,
\begin{align}
\mbox{PSNR}=\log_{10} \frac{\max(Z_o)}{MSE(Z,Z_o)}.
\end{align} 
Both metrics are commonly used in measuring the quality of reconstruction of compressed or noisy signals. 

\begin{figure}[!t]
	\centering
	\setlength\tabcolsep{1.5pt}
	\includegraphics[width=0.9\linewidth]{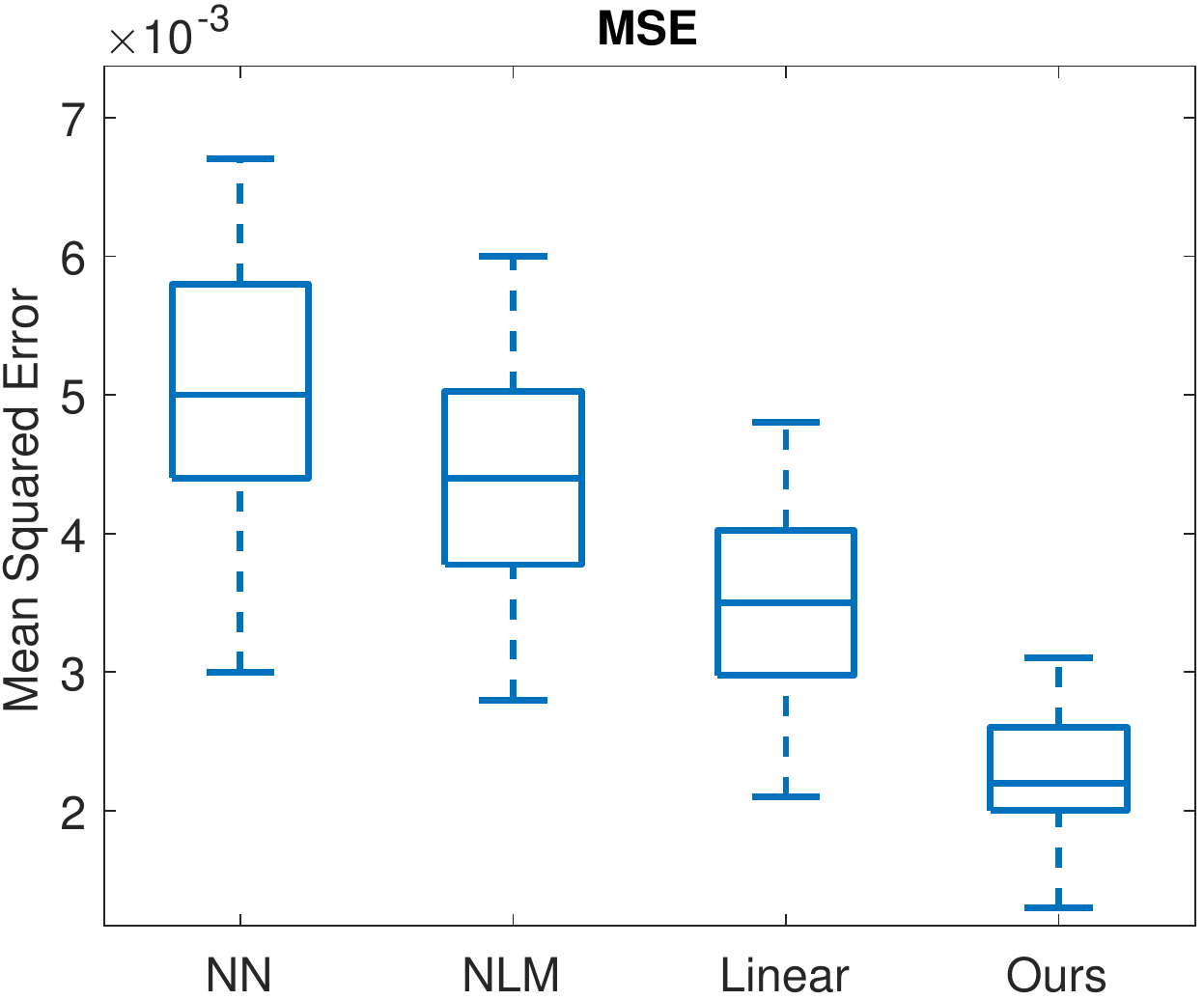}
	\includegraphics[width=0.9\linewidth]{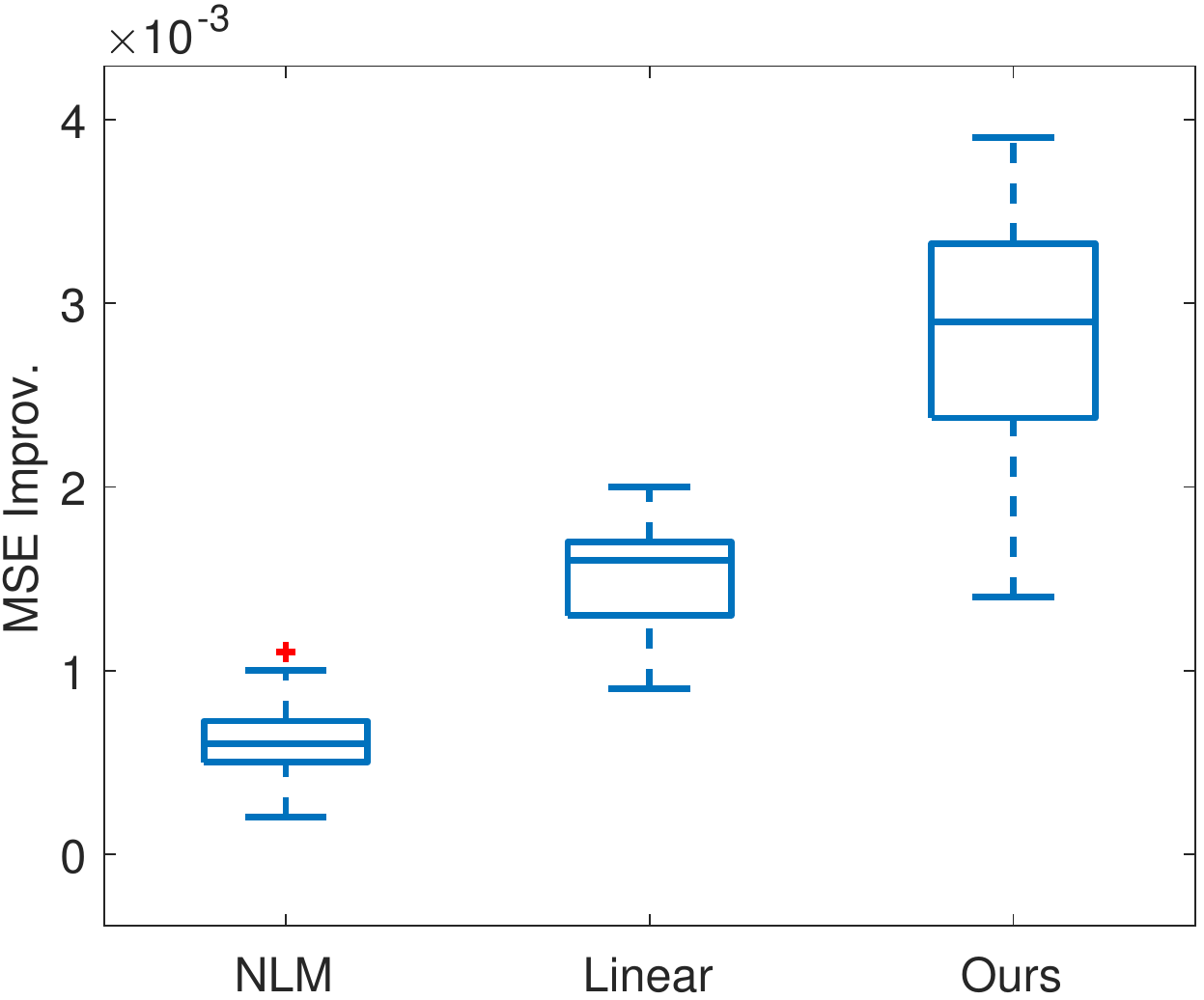}
	\caption{\textbf{Reconstruction accuracy statistics}. Accuracy for different image restoration methods (top), and improvement over nearest neighbor interpolation using MSE (bottom). All statistics were computed over 50 scans randomly chosen from the ADNI dataset. Image intensities are scaled to a~$[0,1]$ range. \vspace{-0.3cm}
	}
	
	\label{fig:local-resultsMetric}
\end{figure}

\subsection{Results}

Fig.~\ref{fig:buckner-images} illustrates representative restored images for subjects in the ADNI dataset. Our method produces more plausible structure. The method restores anatomical structures that are almost entirely missing in the other reconstructions, such as the dura or the sulci {of the temporal lobe} by learning about these structures from the image collection. We provide additional example results in the Supplementary Materials.

Fig.~\ref{fig:local-resultsMetric} reports the error statistics in the ADNI data. Due to high variability of MSE among subject scans, we report improvements of each method over the nearest neighbor interpolation baseline in the same scan. Our algorithm offers significant improvement compared to nearest neighbor, NLM, and linear interpolation ($p \le 10^{-5},10^{-42}$,~$10^{-27}$, respectively). Our method performs significantly better on all subjects. The improvement in MSE is observed in every single scan. Similarly, our method performs consistently better using the PSNR metric (not shown), with mean improvements of up to~$1.4\pm0.44$ compared to the next best restored scan.   

\begin{figure}[t]
	\centering
	\includegraphics[width=0.9\linewidth]{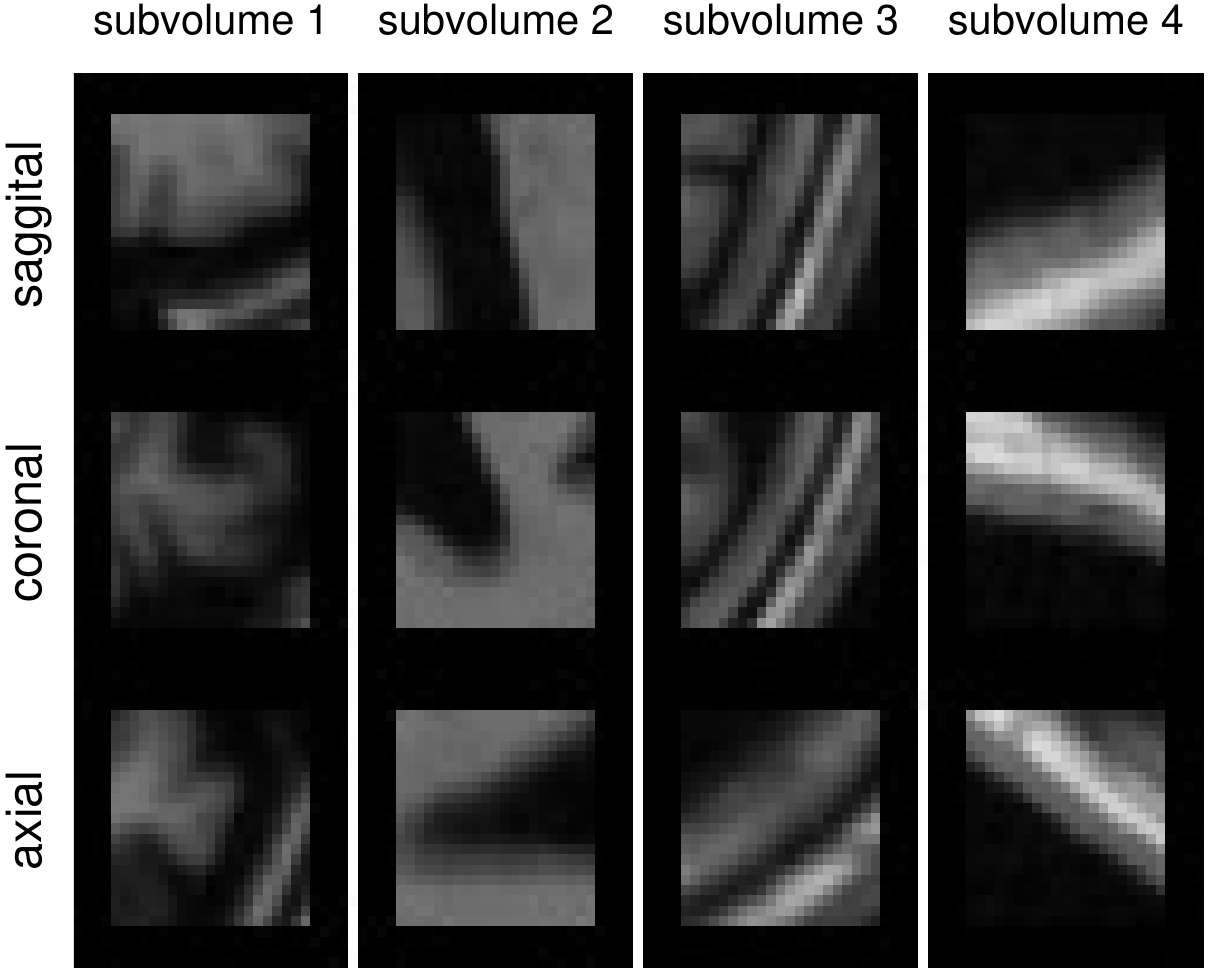}
	\caption{\textbf{Regions used for hyper-parameter analysis}. Representative example of four subvolumes used in analyses, shown in saggital, coronal and axial views.  
	}
	\label{fig:analysis-kd}
\end{figure}

\subsection{Parameter Setting}

We analyze the performance of our algorithm while varying the values of the parameters, and the sparsity patterns of the observed voxels. For these experiments, we use four distinct subvolumes that encompass diverse anatomy from ADNI data, as illustrated in Fig.~\ref{fig:analysis-kd}. We start with isotropic data and use different observation masks as described in each experiment.

\textbf{Hyper-parameters}. We evaluate the sensitivity of our method under different hyper parameters: the number of clusters,~$k \in [1,2,5,10,15]$ and the number of dimensions of the low dimensional embedding~$d \in [10, 20, 30, 40, 50]$. While different regions give optimal results with different settings, overall our algorithm produces comparable results for the middle range of these parameters. We run all of our experiments with~$k=5$ and~$d=30$.

\begin{figure}[t]
	\centering
	\includegraphics[width=0.75\linewidth]{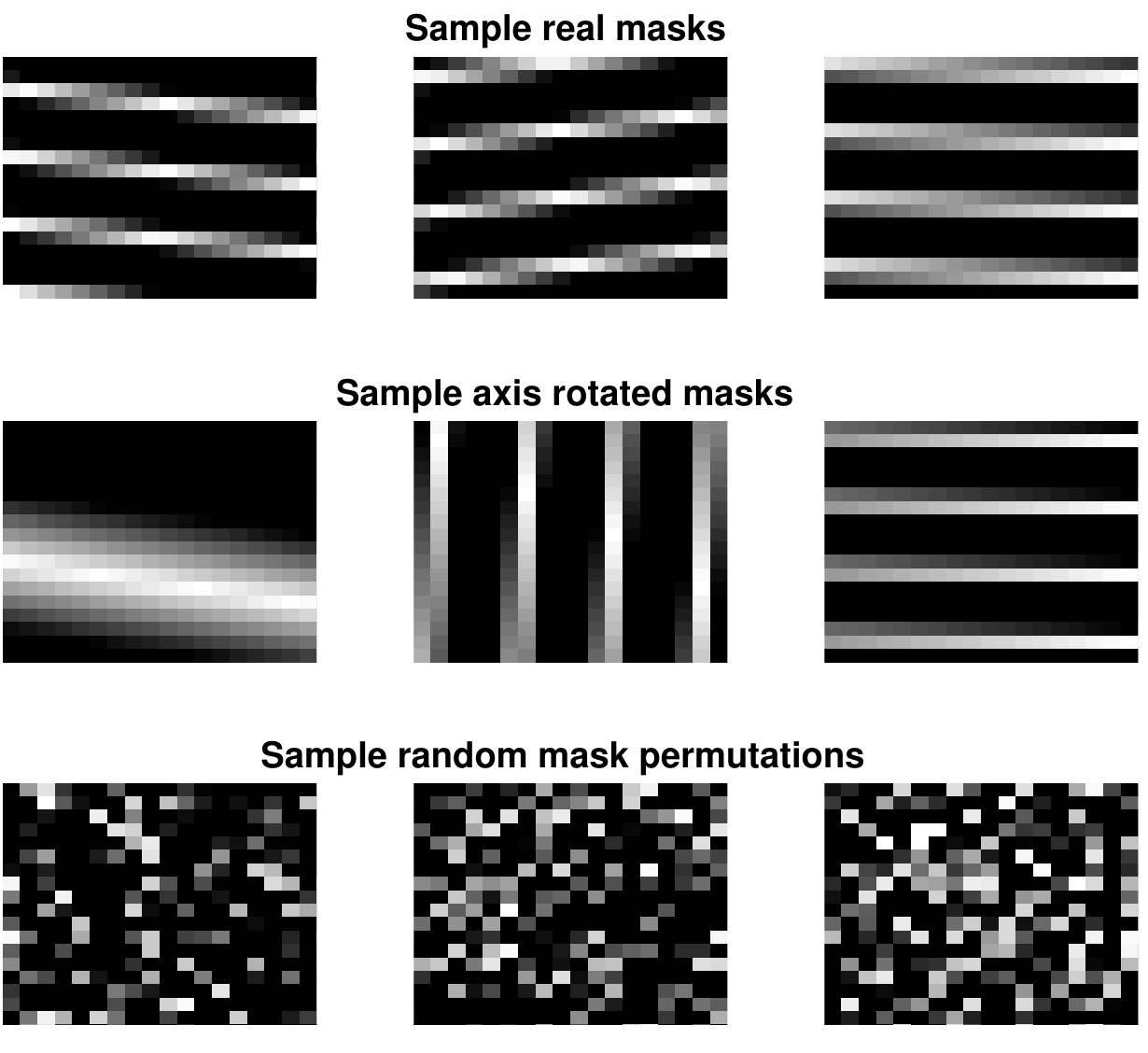}
	
	\vspace{0.5cm}
	
	\includegraphics[width=0.85\linewidth]{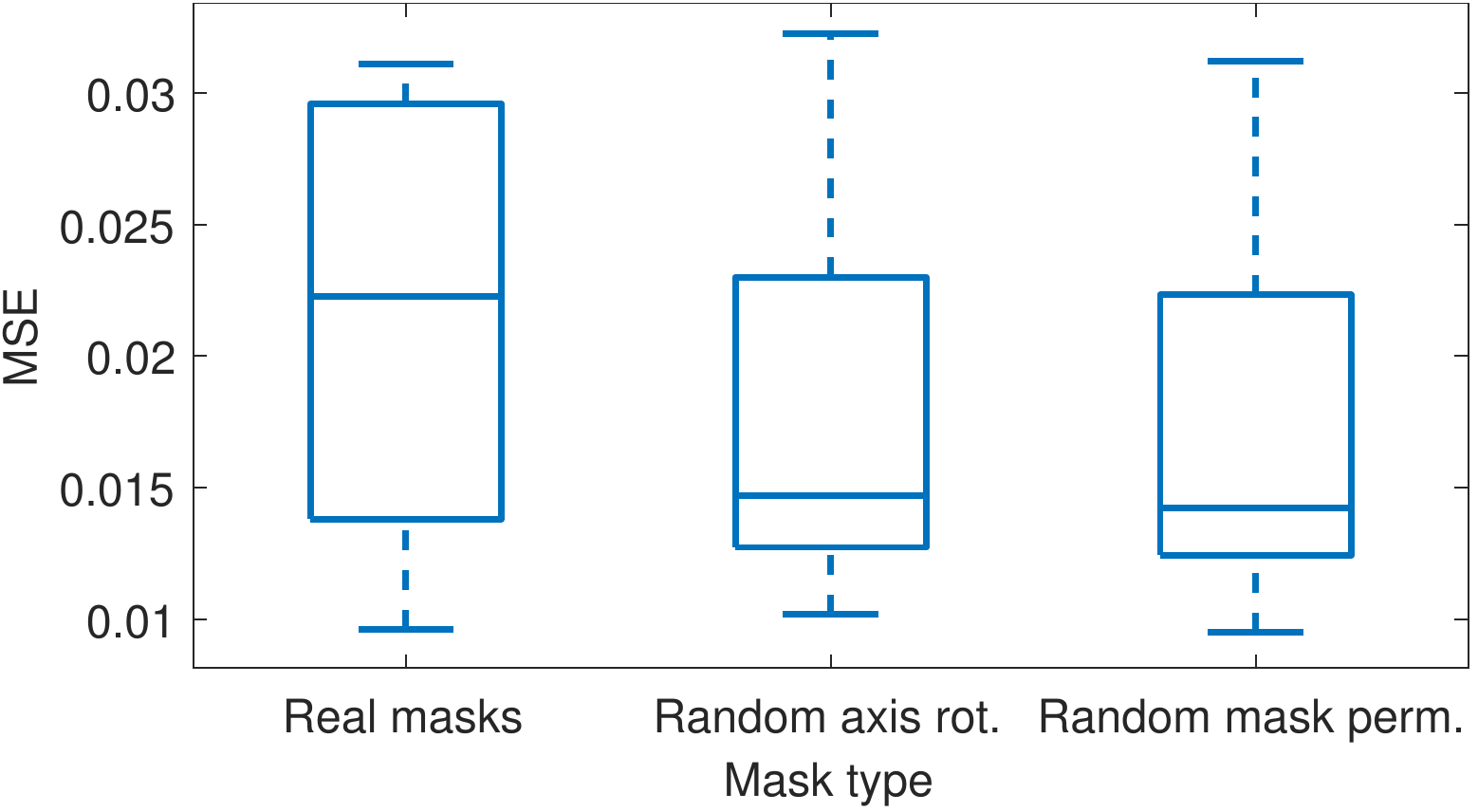}
	\caption{\textbf{Mask Analysis}. Top: example masks for each simulation, shown in the saggital plane. The first experiment reflects the limited variability of axial-only acquisitions, whereas the second and third experiments represent increasingly more varied patterns of observed voxels. Bottom: imputation errors. More varied masks leads to improved reconstructions. \vspace{-0.25cm}}
	\label{fig:analysis-masks}
	%
\end{figure}

\textbf{Sparsity patterns}. First, we evaluate how our algorithm performs under three different mask patterns, all of which allow for the same number of observed voxels. Specifically, we (i) use the true sparsely observed planes as in the first experiment; (ii) simulate random rotations of the observation planes mimicking acquisitions in different directions; and (iii) simulate random mask patterns. The latter setup is useful for denoising or similar tasks, and is instructive of the performance of our algorithm. Fig.~\ref{fig:analysis-masks} demonstrates that our algorithm performs better under acquisition with different directions, and similarly under truly random observations as more entries of the cluster covariance matrices are directly observed.This demonstrates a promising application of this model to other settings where different patterns of image voxels are observed.

\begin{figure}[t]
		\centering
	\includegraphics[width=0.9\linewidth]{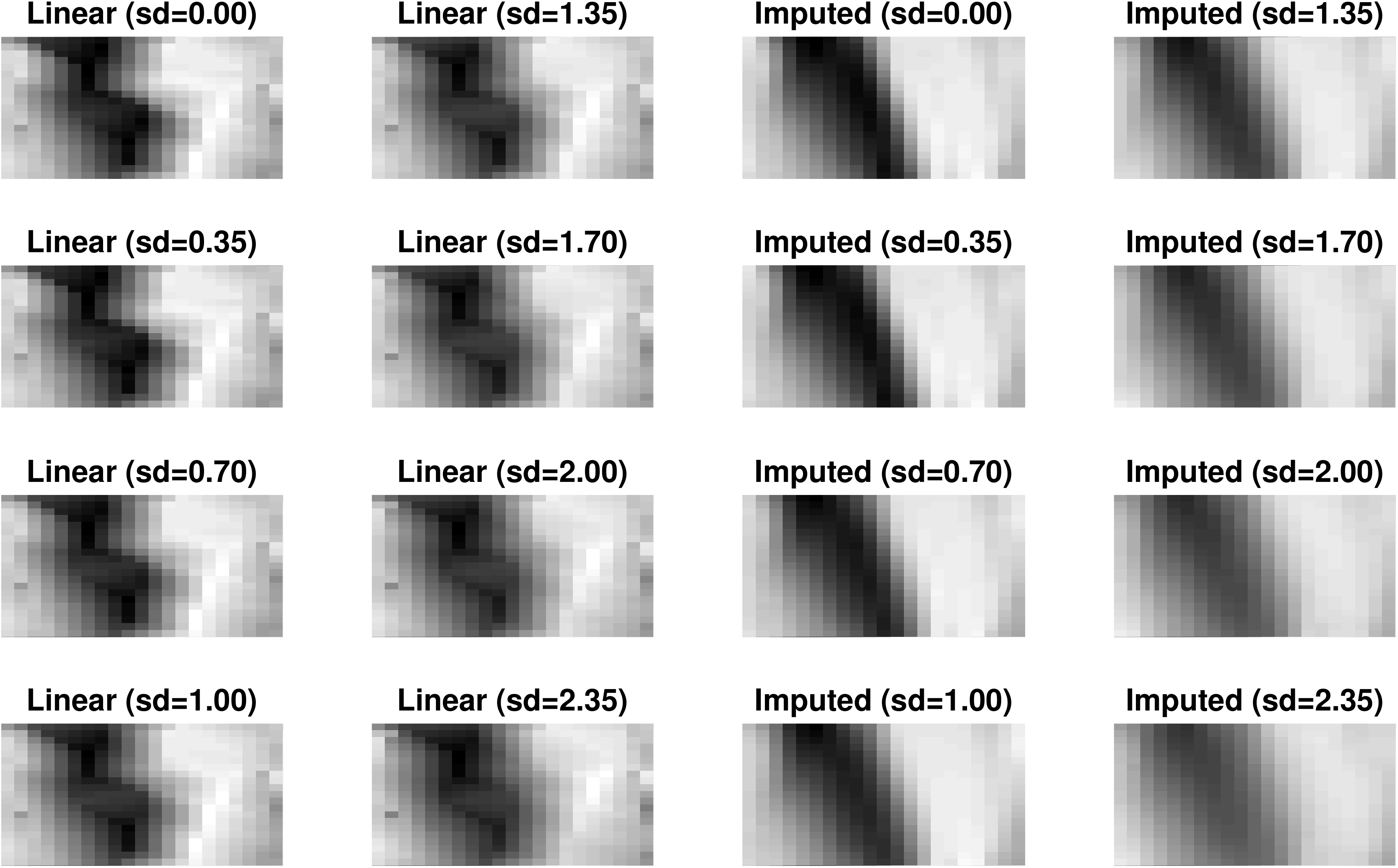}
	
	\vspace{0.5cm}

	\includegraphics[width=0.85\linewidth]{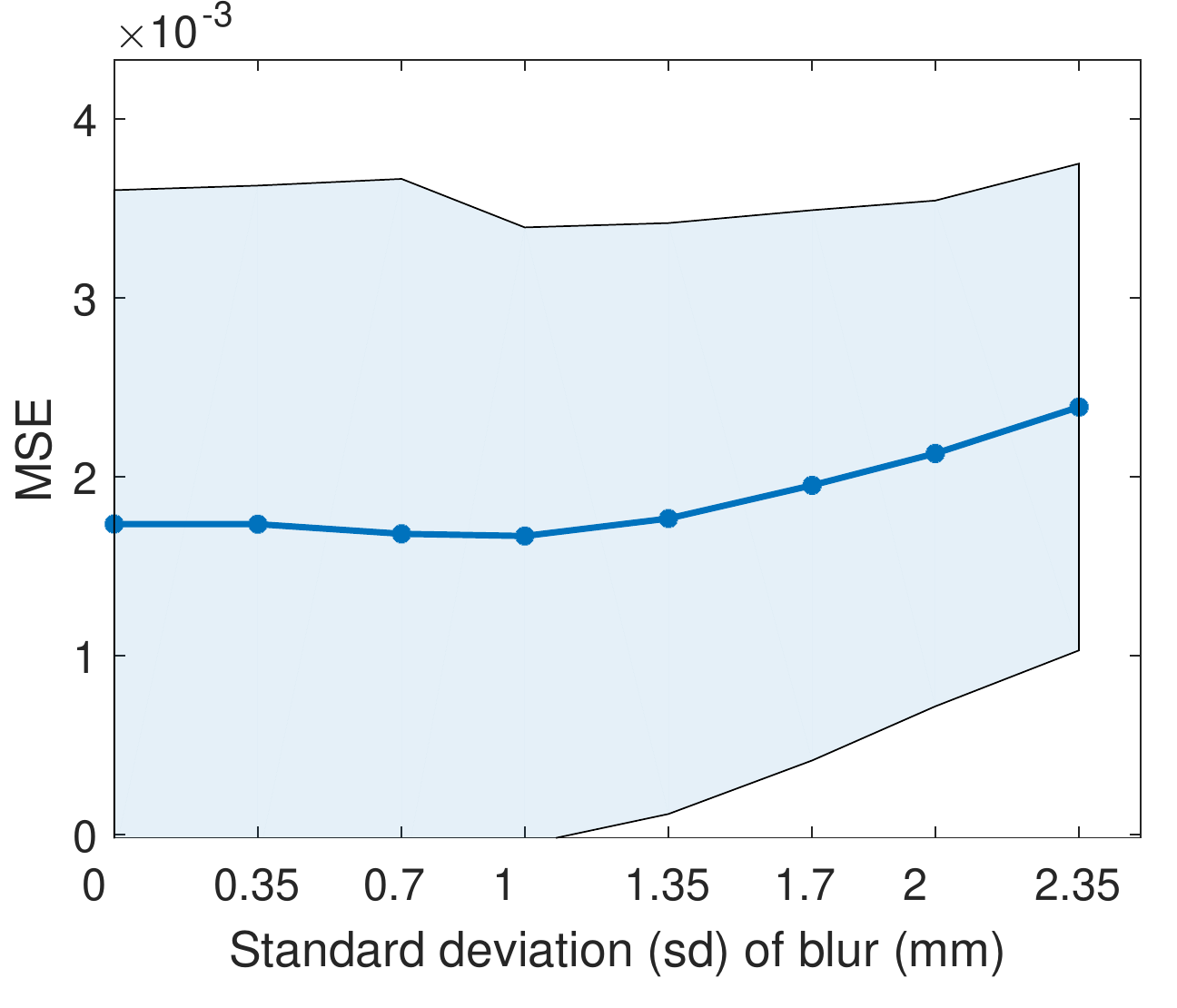}
	\caption{\textbf{Slice thickness simulation}. Top: saggital close-up of a region where axial slices were blurred in the direction perpendicular to the acquisition direction; followed by respective imputed results. Bottom: performance of our algorithm under different slice thickness simulations are shown MSE (solid line) and standard deviation interval (shaded region). \vspace{-0.5cm}}
	\label{fig:analysis-thickness}
	%
	~ 
\end{figure}

\textbf{Slice Thickness}. We also investigate the effects of slice thickness on the results. The model treats the original data as high resolution planes. Here, we simulate varying slice thickness by blurring isotropic data in the direction perpendicular to the slice acquisition direction.  We then use the sampling masks of the scans used in the main experiments to identify observed, albeit blurred, voxels. Fig.~\ref{fig:analysis-thickness} shows that although the algorithm performs worse with larger slice thickness, it provides plausible imputation results. For example, results show minimal noticeable differences, even for a blur kernel of~$\sigma=1$mm, simulating a slice with significant signal contribution from~$4$mm away. Our method, which treats observed slices as thin, is nevertheless robust to slice thicknesses variations.


%
%

\newpage
\subsection{Skull Stripping}

\begin{figure*}[t]
	\centering
	\includegraphics[width=1\linewidth]{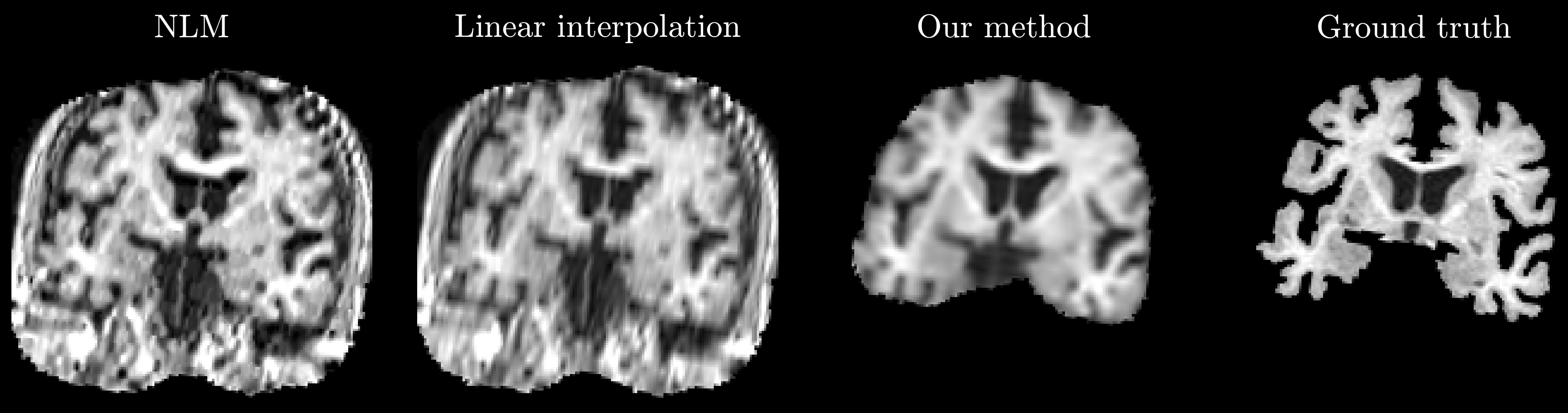}
	\caption{\textbf{Skull Stripping Example}. Example of a skull stripping failure for linear and NLM interpolation. Skull stripping dramatically improves when applied to the imputed image for this example. \vspace{-0.3cm}}
	\label{fig:skull-stripping}
\end{figure*}

We also illustrate how imputed data might facilitate downstream image analysis. Specifically, the first step in many analysis pipelines is brain extraction -- isolating the brain from the rest of the anatomy. Typical algorithms assume that the brain consists of a single connected component separated from the skull and dura by cerebral spinal fluid~\cite{segonne2004}. Thus, they often fail on sparsely sampled scans that no longer include a clear contrast between these regions. Fig.~\ref{fig:skull-stripping} provides an example where the brain extraction fails on the original subject scan but succeeds on our reconstructed image.

\subsection{Clinical Dataset}

%
%
%
We also demonstrate our algorithm on a clinical set of 766 T2-FLAIR brain MR scans in a stroke patient cohort. These scans are severely anisotropic ($0.85\times0.85$mm in-plane, slice separation of~$6$mm). All subjects are affinely registered to an atlas and the intensity is normalized. 


%


Fig.~\ref{fig:stroke-images} illustrates representative restoration improvements in T2-FLAIR scans from a clinical population. Our method produces more plausible structure, as can be especially seen in the close-up panels focusing on anatomical details. {We provide additional example results in the Supplementary Materials.}

\section{Discussion}
\label{sec:discussion}

{\color{blue}
	\textbf{Modeling Choices.} We explicitly model and estimate a latent low-dimensional embedding for each patch. 
	The likelihood model~\eqref{eq:local-gmmppca-likelihood} does not include the latent patch representation~$\bx_i$, leading to observed patch covariance~\mbox{$C_k^{\obsi,\obsi} = W^{\obsi}_k (W^{\obsi}_k)^T + \sigma_k^2 I$}. Since the set of observed voxels~$\mathcal{O}_i$ varies across subjects, the resulting Expectation Maximization algorithm~\cite{dempster1977} becomes intractable if we marginalize the latent representation out before estimation. Introducing the latent structure simplifies the optimization problem. 

	We investigated an alternative modeling choice that instead treats each missing voxel as a latent variable. In particular we consider the missing values of patch~$\by_i$ as latent variables, which can be optionally combined with the latent low-dimensional patch representation. 
	These assumptions lead to an Expectation Conditional Maximization (ECM)~\cite{ilin2010,little2014}, a variant of the Generalized Expectation Maximization where parameter updates depend on the previous parameter estimates. 
	The resulting algorithm estimates the expected missing voxel mean and covariance directly, and then updates the cluster parameters (see Supplementary Materials for a complete derivation).
	The most notable difference between this formulation and simpler algorithms that iteratively fill in missing voxels and then estimate GMM model parameters is in the estimation of the expected data covariance, which captures the covariance of the missing and observed data~(c.f.~\cite{little2014}, Ch.8). We found that compared to the method presented in Section~\ref{sec:mixture-full}, this variant often got stuck in local minima, had difficulty moving away from the initial missing voxel estimates, and was an order of magnitude slower than the presented method. We provide both implementations in our code.
	
	\textbf{Restoration.} 
	Our restoration method assumes that the observed voxels are noisy manifestations of the low dimensional patch representation, and reconstructs the entire patch, including the observed voxels, leading to smoother images. This formulation assumes the original observed voxels are noisy observations of the \textit{true} data. Depending on the downstream analyses, the original voxels could be kept in the reconstruction. In addition, we also investigated an alternative reconstruction method of filling in the missing voxels given the observed voxels as \textit{noiseless} ground truth (not shown).
	This formulation leads to sharper but noisier results. The two restoration methods therefore yield images with different characteristics. This tradeoff is a function of the noise in the original acquisition: higher noise in the clinical acquisition leads to noisier reconstructions using the alternative method, whereas in the ADNI dataset the two methods perform similarly.  In addition, imputation can be achieved by sampling the posterior distribution rather than using conditional mean estimation, enabling a better estimate of the residual noise for downstream analysis.

	\begin{figure*}[!t]
		\centering
		\includegraphics[width=1\linewidth]{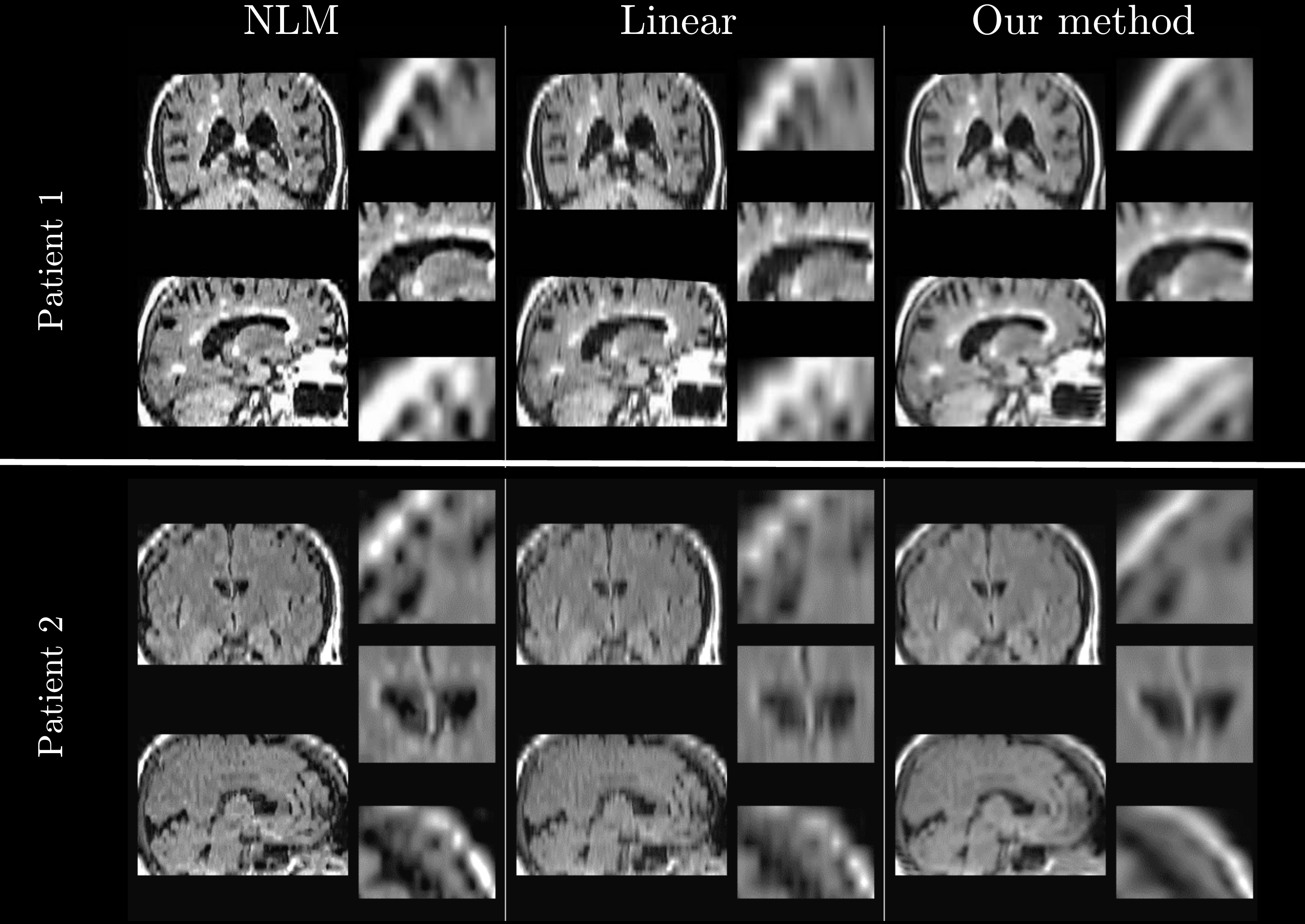}
		\caption{\textbf{Representative restorations in the clinical dataset}. Reconstruction using NLM, linear interpolation and our method for two representative subjects. Our method reconstructs more plausible substructures, as can be especially seen in the close-up panels of the skull and  the periventricular region. Additional examples are available in the Supplementary Materials. \vspace{-0.4cm}}
		\label{fig:stroke-images}
	\end{figure*}

	\textbf{Usability.} Our model assumes that whether a voxel is observed is independent of the intensity of that voxel. Although the voxels missing in the sparsely sampled images clearly form a spatial pattern, we assume there is no correlation with the actual intensity of the voxels. The model can therefore be learned from data with varying sparseness patterns, including restoring data in all acquisition directions simultaneously.
	
	The proposed method can be used for general image imputation using datasets of varying resolution. For example, although acquiring a large high resolution dataset for a clinical study is often infeasible, our algorithm will naturally make use of any additional image data available. Even a small number of acquisitions in different directions or higher resolution than the study scans promise to improve the accuracy of the resulting reconstruction.
	
	The presented model depends on the image collection containing similar anatomical structures roughly aligned, such as affinely aligned brain or cardiac MR scans. Smaller datasets that contain vastly different scans, such as traumatic brain injuries or tumors, may not contain enough consistency to enable the model to learn meaningful covariation. However, a wide range of clinical datasets contain the anatomical consistency required, and can benefit from the proposed method.

	\textbf{Initialization.} We experimented with several initialization schemes, and provide them in our implementation. A natural initialization is to first learn a simple GMM from the linearly interpolated volumes, and use the resulting parameter values as initializations for our method. This leads to results that improve on the linear interpolation but still maintain somewhat blocky effects caused by interpolation. More agnostic initializations, such as random parameter values, lead to more realistic anatomy but noisier final estimates. Different methods perform well in different regions of the brain. The experimental results are initialized by first learning a simple GMM from the linearly interpolated volumes, and using the resulting means with diagonal covariances as an initial setting of the parameters. We start with a low dimensional representation to be of dimension~$1$, and grow it with every iteration up to the desired dimension. We found that this approach outperforms all other strategies. 
	
}

\section{Conclusions}

We propose an image imputation method that employs a large collection of low-resolution images to infer fine-scale anatomy of a particular subject.  We introduce a model that captures anatomical similarity across subjects in large clinical image collections, and imputes, or fills in, the missing data in low resolution scans.  
The method produces anatomically plausible volumetric images consistent with sparsely sampled input scans.

Our approach does not require high resolution scans or expert annotations for training. 
We demonstrate that our algorithm is robust to many data variations, including varying slice thickness. The resulting method enables the use of untapped clinical data for large scale scientific studies and promises to facilitate novel clinical analyses.

\newpage
\section*{Acknowledgment}
We acknowledge the following funding sources: NIH NINDS R01NS086905, NIH NICHD U01HD087211, NIH NIBIB NAC P41EB015902, NIH R41AG052246-01, 1K25EB013649-01, 1R21AG050122-01, NSF IIS 1447473, Wistron Corporation, and SIP. 

Data collection and sharing for this project was funded by the Alzheimer's Disease
Neuroimaging Initiative (ADNI) (National Institutes of Health Grant U01 AG024904) and
DOD ADNI (Department of Defense award number W81XWH-12-2-0012). ADNI is funded
by the National Institute on Aging, the National Institute of Biomedical Imaging and
Bioengineering, and through generous contributions from several agencies listed at~\url{http://adni.loni.usc.edu/about/}.

\appendices

%
%
%


\clearpage 
\section{Expectation Maximization Updates}
\label{app:em}

Following~\eqref{eq:local-gmmppca-likelihood}, the complete likelihood of our model is:
\begin{align}
p(\mathcal{Y}^{\obs}, \mathcal{X}; \theta) &= \prod_i \sum_k \pi_k \mathcal{N}(\by_{i}^{\obsi}, \bx_i; \bmu_{k}^{\obsi}, \bC_{k}^{\obsi\obsi}),
\label{eq:app-local-gmmppca-likelihood}
\end{align}
where~$\mathcal{X} = \{x_i\}$. The expectation of this probability is then
\begin{align}
Q(\btheta|\btheta) &= \Expect_{\mathcal{X} | \mathcal{Y}^{\obs}, \btheta} \left[ \log p(\mathcal{Y}^{\obs}, \mathcal{X}; \theta) \right] \nonumber \\
&= \sum_{i,k} \Expect [ k ( -\frac{d}{2} \log 2\pi - \frac{1}{2} \log|\sigma^2_k \bI| \nonumber \\
&   - \frac{1}{2\sigma^2_k}  (\by_{i}^{\obsi} - \bW^{\obs}_k\bx_i - \bmu_k^{\obsi})^T (\by_{i}^{\obsi} - \bW^{\obs}_k\bx_i - \bmu_k^{\obsi}) \nonumber \\
&   - \frac{d}{2} \log 2\pi - \frac{1}{2} \log| \bI| - \frac{1}{2} \bx_i\bx_i^T ) ]. \label{app:exp-compl-lik}
\end{align}
Computing this expectation requires evaluating~$\Expect[k]$,~$\Expect[\bx_i|k]$, and~$\Expect[\bx_i\bx^T_i|k]$, which is trivially done to obtain the expectation step updates~\eqref{eq:local-gmmppca-estep-mem} -\eqref{eq:local-gmmppca-estep-S} .

For the maximization step, we optimize~\eqref{app:exp-compl-lik} with respect to the model parameters.
\begin{align}
\frac{\partial Q}{\partial \bmu_k} &\propto \sum_{i\in\mathcal{P}_j} \frac{\partial}{\partial \bmu_k}  \Expect  \left[k (\by_i^{\obsi} - \bW_k^{\obsi}\bx_i - \bmu_k^{\obsi})^T (\by_i^{\obsi} - \bW_k^{\obsi}\bx_i - \bmu_k^{\obsi}) \right] \nonumber  \\
&\propto \sum_{i\in\mathcal{P}_j}   \Expect \left[ k (\by_i^{\obsi} - \bW_k^{\obsi}\bx_i - \bmu_k^{\obsi}) \right] \nonumber  \\
&= \sum_{i\in\mathcal{P}_j} \gamma_{ik} (\by_i^{\obsi} - \bW_k^{\obsi}\widehat{\bx}_{ik} - \bmu_k^{\obsi}) = 0, \nonumber 
\end{align}
\begin{align}
\mu_k^j &= \frac{1}{\sum_{i'} \gamma_{i'k}} \sum_{i\in\mathcal{P}_j} \gamma_{ik}(y_i^j - \bW_k^{j}\widehat{\bx}_{ik})
\label{eq-app-local:ppca-mstep-mu}
\end{align}
\begin{align}
\frac{\partial Q}{\partial \bW_k} &\propto \sum_{i\in\mathcal{P}_j} \frac{\partial  \Expect \left[k(\by_i^{\obsi} - \bW_k^{\obsi}\bx_i - \bmu_k^{\obsi})^T (\by_i^{\obsi} - \bW_k^{\obsi}\bx_i - \bmu_k^{\obsi}) \right]}{\partial \bW_k} \nonumber  \\
&\propto \sum_{i\in\mathcal{P}_j} \Expect \left[ k (\bW_k^{\obsi}\bx_i - (\by_i^{\obsi}- \bmu_k^{\obsi})) \bx_i^T \right] \nonumber  \\
&= \sum_{i\in\mathcal{P}_j} \gamma_{ik} \bW^{\obsi} (\widehat{\bx}_{ik} \widehat{\bx}_{ik}^{T} + \bS_{ik}) - (\by_i^{\obsi}- (\bmu_k^{\obsi})) \widehat{\bx}_{ik}^{T} = 0 \nonumber  \\
\bW_k^j &= \left[ \sum_{i\in\mathcal{P}_j} \gamma_{ik} (\widehat{\bx}_{ik} \widehat{\bx}_{ik}^{T} + \bS_{ik}) \right]^{-1} \sum_{i\in\mathcal{P}_j} \gamma_{ik} (y_i^j- \mu_k^j) \widehat{\bx}_{ik}^{T} \nonumber \\
&= \sum_{i\in\obsj} \delta_{ik} (y_{i}^{j} - \mu_{k}^{j}) \widehat{\bx}_{ik}^{T} A_j^{-1}
\label{eq-app-local:ppca-mstep-W}
\end{align}
where~$\delta$ and~$A$ are defined in~\eqref{eq:mean-exp-delta} and~\eqref{eq:mean-exp-cov}, respectively. By combining~\eqref{eq-app-local:ppca-mstep-W} and~\eqref{eq-app-local:ppca-mstep-mu}, we obtain
\begin{align}
\mu_k^j \sum_{i} \gamma_{ik} &= \sum_{i\in\mathcal{P}_j} \gamma_{ik}y_i^j - \bW_k^{j} \sum_{i\in\mathcal{P}_j} \gamma_{ik} \widehat{\bx}_{ik} \nonumber \\
\mu_k^j  &= \sum_{i\in\mathcal{P}_j} \delta_{ik}y_i^j - \bW_k^{j} \bb_j \nonumber \\
\mu_k^j  &= \sum_{i\in\mathcal{P}_j} \delta_{ik}y_i^j - \sum_{i\in\obsj} \delta_{ik} (y_{i}^{j} - \mu_{k}^{j}) \widehat{\bx}_{ik}^{T} A_j^{-1} \bb_j \nonumber 
\end{align}
\begin{align}
\mu_k^j(1 &- \sum_{i\in\obsj} \delta_{ik} \widehat{\bx}_{ik}^{T} A_j^{-1}\bb_j)  = \sum_{i\in\mathcal{P}_j} \delta_{ik}y_i^j(1 - \widehat{\bx}_{ik}^{T} A_j^{-1}\bb_j)\nonumber \\
\mu_k^j  &= \frac{\sum_{i\in\mathcal{P}_j} \delta_{ik}y_i^j(1 - \widehat{\bx}_{ik}^{T} A_j^{-1}\bb_j)}{\sum_{i\in\obsj}\delta_{ik} (1 - \widehat{\bx}_{ik}^{T} A_j^{-1}\bb_j)}.
\label{app:mstep-mu-final}
\end{align}
We therefore update~$\mu_k^j$ via~\eqref{app:mstep-mu-final}, followed by~$\bW_k^j$ using~\eqref{eq-app-local:ppca-mstep-W}. Finally,
\begin{align}
\frac{\partial Q}{\partial \sigma^2_k} &\propto \sum_{i\in\mathcal{P}_j} - \frac{\partial}{\partial \sigma^2_k} \frac{1}{\sigma^2_k}  \nonumber\\ &~\quad \Expect \left[k (\by_i^{\obsi} - \bW_k^{\obsi}\bx_i - \bmu_k^{\obsi})^T (\by_i^{\obsi} - \bW_k^{\obsi}\bx_i - \bmu_k^{\obsi}) \right] \nonumber \\
&- \frac{\partial}{\partial \sigma^2_k} N\log\sigma^2_k \nonumber  \\
&= \sum_{i\in\mathcal{P}_j} \frac{1}{\sigma^4_k} \Expect \left[k (\by_i^{\obsi} - \bW_k^{\obsi}\bx_i - \bmu_k^{\obsi})^T (\by_i^{\obsi} - \bW_k^{\obsi}\bx_i - \bmu_k^{\obsi}) \right] \nonumber \\ &-\frac{N}{\sigma^2_k} = 0 \nonumber  
\end{align}
\begin{align}
\sigma^{2}_k &= \sum_{i\in\mathcal{P}_j} \Expect \left[k (\by_i^{\obsi} - \bW_k^{\obsi}\bx_i - \bmu_k^{\obsi})^T (\by_i^{\obsi} - \bW_k^{\obsi}\bx_i - \bmu_k^{\obsi}) \right] \nonumber \\
\sigma_k^{2} &= \frac{\sum_j\sum_{i\in\obsj} \gamma_{ik} \left[ (y_{i}^{j} - \mu_{k}^{j} - \bW_{k}^j\widehat{\bx}_{ik})^2 + \bW_{k}^j \bS_{ik} (\bW_{k}^j)^{T} \right]}{\sum_j\sum_{i\in\obsj} \gamma_{ik}}.
\label{eq-app-local:ppca-mstep-sigma}
\end{align}


\IEEEtriggeratref{25}

\ifCLASSOPTIONcaptionsoff
  \newpage
\fi

\scriptsize
\bibliography{bibliography}
\bibliographystyle{plain}

\clearpage
\begin{center}
	{\Large SUPPLEMENTARY MATERIAL}
\end{center}
\section*{Derivation of Alternative Model}
\label{app:alternate}

\normalsize
In this section, we explore the parameter estimation for an alternative model. Specifically, letting~$\mathcal{M}_i$ be the set of missing voxels of patch~$y_i$, we treat~$\by_i^{\misi}$ as latent variables, instead of explicitly modeling a low-dimensional representation~$\bx$.
We show the maximum likelihood updates of the model parameters under the likelihood~\eqref{eq:local-gmmppca-likelihood}. We employ the Expectation Conditional Maximization (ECM)~\cite{ilin2010,little2014} variant of the Generalized Expectation Maximization, where parameter updates depend on the previous parameter estimates. 

The complete data likelihood is
\begin{align}
p(\mathcal{Y}; \theta) &= \prod_i \sum_k \pi_k \mathcal{N}(\by_{i}^{\obsi}, \by_{i}^{\misi}; \bmu_{k}^{\obsi}, \bSigma_{k}^{\obsi\obsi}).
\label{eq:app-local-gmmml-likelihood}
\end{align}

The \textbf{expectation step} updates the statistics of the missing data, computed based on covariates of the known and unknown voxels: 
\begin{align}
\gamma_{ik} &\equiv  \Expect[k_i] \nonumber \\ &=
 \frac{\pi_k \mathcal{N}(\by_{i}^{\obsi}; \bmu_{k}^{\obsi},\bSigma_{k}^{\obsi})}{\sum_k \pi_k \mathcal{N}(\by_{i}^{\obsi}; \bmu_{k}^{\obsi}, \bSigma_{k}^{\obsi} )}  \label{eq:app:local-gmmpcar-estep-k}  \\
\widehat{y}_{ij} &\equiv \Expect[y_{ij}]  \nonumber \\ 
&= \left\{
\begin{array}{ll}
{y}_{ij}  & \parbox{2cm}{if ${y}_{ij}$ \\ is observed} \\
{\mu}_{ij} + {\bSigma}^{j\obsi}_{i} ({\bSigma}^{\obsi\obsi}_{i})^{-1} ({\by}^{\obsi}_{i} - {\bmu}^{\obsi}) & \mbox{otherwise}  %
\end{array}
\right.
\label{eq:app:local-gmmpcar-estep-y} \\
\widehat{s}_{ijl} &\equiv \Expect \left[ y_{ij} y_{il} \right] - \Expect[y_{ij}] \Expect[y_{il}] \nonumber \\ 
&=
\left\{
\begin{array}{ll}
0  & \parbox{2cm}{if ${y}_{ij}$ or ${y}_{il}$ \\ is observed} \\
{\bSigma}^{jl}_{i} - ({\bSigma}^{\obsi j}_{i})^T({\bSigma}^{\obsi\obsi}_{i})^{-1}{\bSigma}^{\obsi l}_{i} \, \, \: \quad & \mbox{otherwise}  %
\end{array}
\right.
\label{eq:app:local-gmmpcar-estep-c}
\end{align}
where the correction in~${\widehat{s}}_{ijl}$ can be interpreted as the uncertainty in the covariance estimation due to the missing values. 

\newpage
\vspace*{0.9cm}
Given estimates for the missing data, the \textbf{maximization step} leads to familiar Gaussian Mixture Model parameters updates:
\begin{align}
\bmu_k &=  \frac{1}{\gamma_{ik}} \sum_i   \gamma_{ik} {\widehat{\by}}_{ik}  \label{eq:app:local-gmmpcar-mstep-gmm-k} \\
\bSigma_k	&= \frac{1}{\gamma_{ik}} \sum_i  \gamma_{ik} \left[ (\widehat{\by}_{ik}-\bmu_k) (\widehat{\by}_{ik} - \bmu_k)^T + {\bS}_i^T \right]. \label{eq:app:gmmcar-mstep-Sigma} \\
\bpi_k &=  \frac{1}{N} \sum_i   \gamma_{ik}  
\label{eq:app:local-gmmpcar-mstep-gmm}
\end{align}
where~$\left[{\bS}_i\right]_{jl} = \widehat{s}_{ijl}$.

In additional to the latent missing voxels, we can still model each patch as coming from a low dimensional representation. We form~$\bC_k = \bW_k \bW_k^{T} + \sigma_k^{2} \bI$ as in~\eqref{eq:gmm-pca}, leading to the complete data likelihood:
\begin{align}
p(\mathcal{Y}; \theta) &= \prod_i \sum_k \pi_k \mathcal{N}(\by_{i}^{\obsi}, \by_{i}^{\misi}; \bmu_{k}^{\obsi}, \bC_{k}^{\obsi\obsi}).
\label{eq:app-local-gmmml-likelihood}
\end{align}
The \textbf{expectation steps} are then unchanged from~\eqref{eq:app:local-gmmpcar-estep-k}-\eqref{eq:app:local-gmmpcar-estep-c} with~$\bC_k$ replacing~$\bSigma_k$. The \textbf{maximization steps} are unchanged from~\eqref{eq:app:local-gmmpcar-mstep-gmm-k}-\eqref{eq:app:local-gmmpcar-mstep-gmm}, with~$\bSigma_k$ now the \textit{empirical} covariance in~\eqref{eq:app:gmmcar-mstep-Sigma}.
%
We let~$\bU \bLambda \bV^T = \mbox{SVD}(\bSigma_k)$ be the singular value decomposition of~$\bSigma_k$, leading to the low dimensional updates
\begin{align}
\sigma_k^{2} &\leftarrow \frac{1}{d-q} \sum_{j=d+1}^d \bLambda(j,j) \\
\bW_k &\leftarrow \bU (\bLambda - \sigma^2 \bI)^{1/2}.
\label{eq:app:local-gmmpcar-mstep-pca}
\end{align}
Finally, we let~$\bC_k = \bW_k \bW_k^{T} + \sigma_k^{2} \bI$. 

Unfortunately, both learning procedures involve estimating all of the missing voxel covariances, leading to a large and unstable optimization.

\begin{figure*}[!p]
	\centering
	\includegraphics[width=0.8\linewidth]{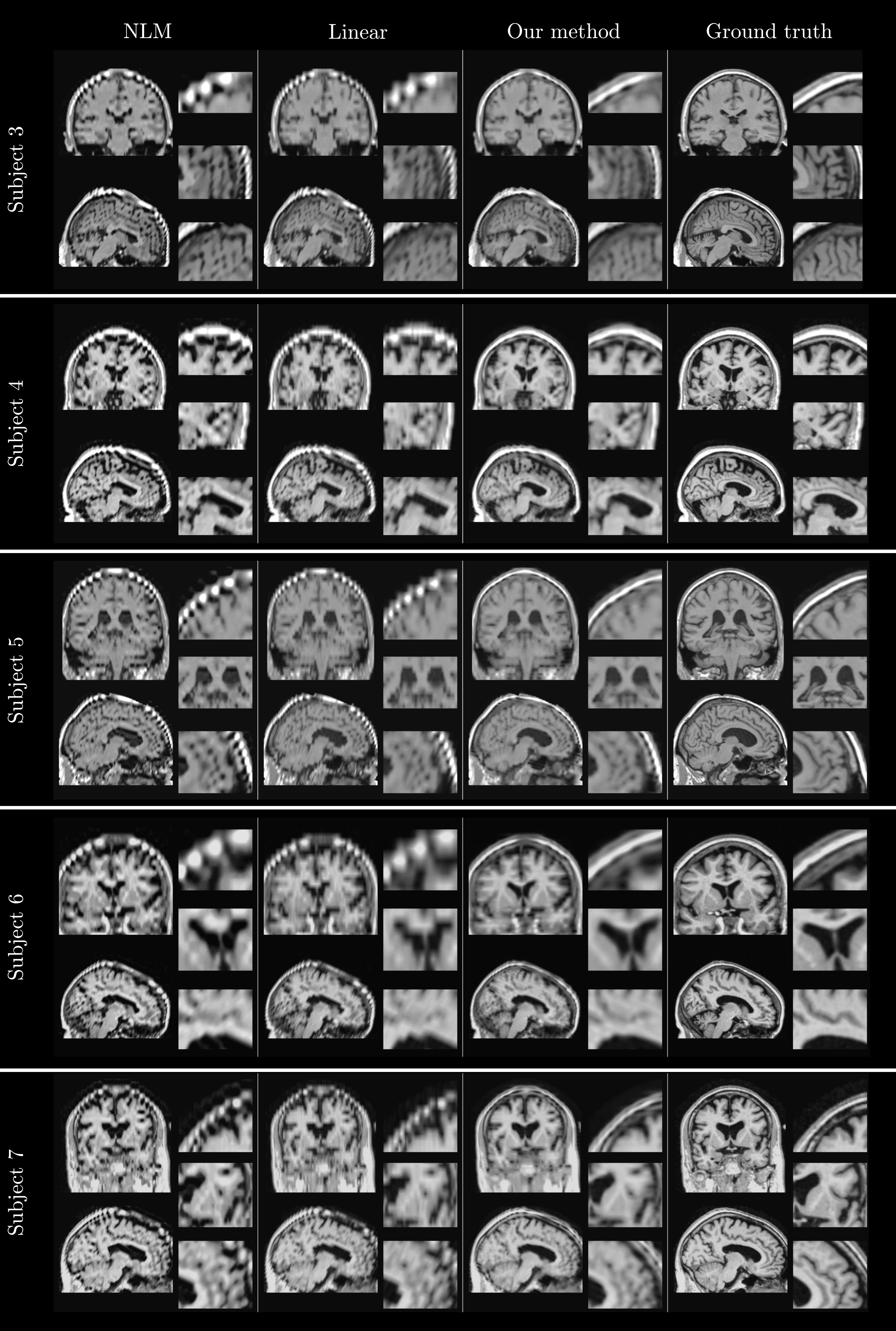}
	\caption{\textbf{Additional restorations in the ADNI dataset}. Reconstruction by NLM, linear interpolation, and our method, and the original high resolution images. }
	\label{fig:buckner-images-sup}
\end{figure*}

\begin{figure*}[!t]
	\centering
	\includegraphics[width=0.7\linewidth]{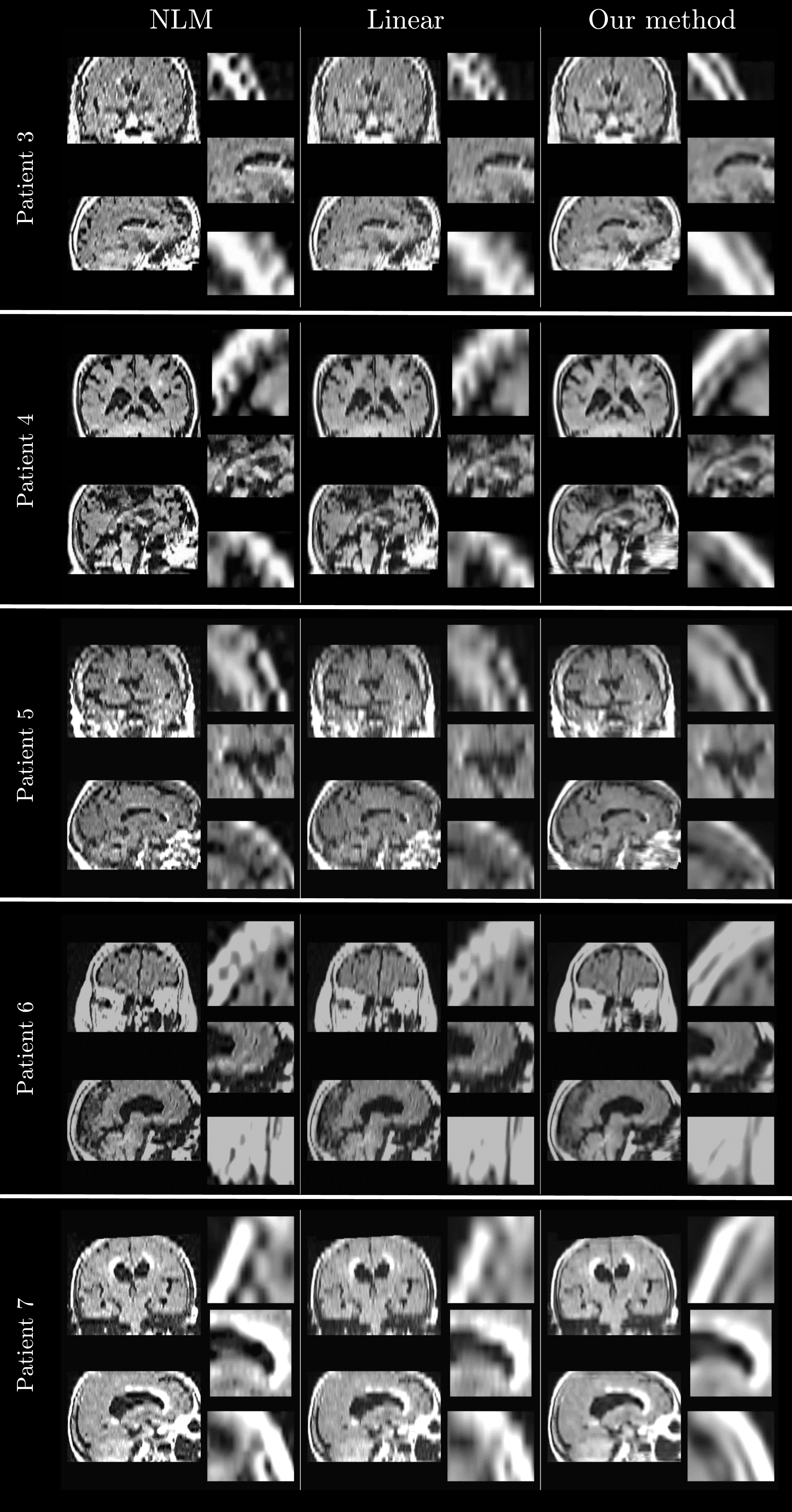}
	\caption{\textbf{Additional restorations in the clinical dataset}. Reconstruction using NLM, linear interpolation and our method.}
	\label{fig:stroke-images-sup}
\end{figure*}

\end{document}